%% file: main.tex
\pdfoutput=1

\documentclass[11pt]{article}

\usepackage{acl}

\usepackage{times}
\usepackage{latexsym}

\usepackage[T1]{fontenc}

\usepackage[utf8]{inputenc}

\usepackage{microtype}

\usepackage{amsmath}
\usepackage{hyperref}
\usepackage{url}
\input{files/macro}
\input{files/math_commands}

\usepackage{todonotes}
\usepackage{pifont}
\usepackage{graphicx}
\usepackage{amssymb}
\usepackage{comment}
\usepackage{enumitem}
\usepackage{caption}
\usepackage{subcaption}
\usepackage{booktabs}
\usepackage{tikz}
\usetikzlibrary{trees}

\usepackage{subfiles}
\usepackage[export]{adjustbox}
\usepackage{comment}
\usepackage{wrapfig}

\usepackage{tabu}
\usepackage{algorithm}
\usepackage[noend]{algpseudocode}

\definecolor{commentcolor}{RGB}{110,154,155}   

\usepackage{color, colortbl}
\definecolor{LRed}{rgb}{1,.9,.9}
\definecolor{LGreen}{rgb}{.9,1,0.9}
\definecolor{LBlue}{rgb}{.9,.9,1}
\definecolor{LYellow}{rgb}{1,1,0.9}

\definecolor{lightblue}{rgb}{0.68, 0.85, 0.9}
\definecolor{lavender}{rgb}{0.9, 0.9, 0.98}
\definecolor{lightyellow}{rgb}{1.0, 1.0, 0.88}
\definecolor{magicmint}{rgb}{0.67, 0.94, 0.82}
\definecolor{palepink}{rgb}{0.98, 0.85, 0.87}
\definecolor{bubbles}{rgb}{0.91, 1.0, 1.0}

\title{Rethinking Self-Supervision Objectives for \\ Generalizable Coherence Modeling}

\author{Prathyusha Jwalapuram$^\dagger$ \and Shafiq Joty$^\dagger$$^\S$ \and Xiang Lin$^\dagger$\\
  $^\dagger$Nanyang Technological University, Singapore \\
  $^\S$Salesforce Research Asia, Singapore \\
  {\tt\{jwal0001,srjoty,linx0057\}@ntu.edu.sg}
}

\begin{document}
\maketitle
\begin{abstract}

{Given the claims of improved text generation quality across various pre-trained neural models, we consider the coherence evaluation of machine generated text to be one of the principal applications of coherence models that needs to be investigated. Prior work in neural coherence modeling has primarily focused on devising new architectures for solving the permuted document task. We instead use a basic model architecture and show significant improvements over state of the art within the same training regime. We then design a harder self-supervision objective by increasing the ratio of negative samples within a contrastive learning setup, and  enhance the model further through automatic hard negative mining coupled with a large global negative queue encoded by a momentum encoder. We show empirically that increasing the density of negative samples improves the basic model, and using a global negative queue further improves and stabilizes the model while training with hard negative samples. We evaluate the coherence model on task-independent test sets that resemble {real-world applications} and show significant improvements in coherence evaluations of downstream tasks.\footnote{Our code and data are available at  \href{https://ntunlpsg.github.io/project/coherence-paradigm}{https://ntunlpsg.github.io/project/coherence-paradigm}}} 


\end{abstract}

\section{Introduction}
\label{sec:intro}
\input{files/introduction}

\section{Datasets}
\label{data}

\input{files/datasets}

\section{Methodology}
\label{sec:exps}
\input{files/experiments}

\section{Analysis}
\label{sec:analysis}
\input{files/analysis}

\section{Conclusion}
\label{sec:concl}
\input{files/conclusions}

\section*{Acknowledgements}
{We would like to thank the Senior Area Chairs of ACL 2022 for evaluating our paper on its merits, and the reviewers and meta-reviewer of ARR for their reviews. We would also like to thank our colleagues Mathieu Ravaut and Han Cheol Moon for their valuable inputs.}

\section*{Ethics Statement}
\label{sec:ethics}
\input{files/ethics}

\bibliography{anthology,references}
\bibliographystyle{acl_natbib}

\newpage
\appendix

\section{Appendix}

\label{sec:appendix}

\input{files/appendix}

\end{document}

%% file: files/macro.tex
\usepackage{xspace}

\usepackage{booktabs,arydshln}

\usepackage[nameinlink]{cleveref}
\crefformat{section}{\S#2#1#3} 
\crefformat{subsection}{\S#2#1#3}
\crefformat{subsubsection}{\S#2#1#3}

\def\adl@drawiv#1#2#3{%
        \hskip.5\tabcolsep
        \xleaders#3{#2.5\@tempdimb #1{1}#2.5\@tempdimb}%
                #2\z@ plus1fil minus1fil\relax
        \hskip.5\tabcolsep}
\newcommand{\cdashlinelr}[1]{%
  \noalign{\vskip\aboverulesep
           \global\let\@dashdrawstore\adl@draw
           \global\let\adl@draw\adl@drawiv}
  \cdashline{#1}
  \noalign{\global\let\adl@draw\@dashdrawstore
           \vskip\belowrulesep}}

\newcommand{\blue}[1]{\textcolor{blue}{#1}}

\newcommand{\cmark}{\ding{52}}%
\newcommand{\xmark}{\ding{56}}

\newcommand{\storycloze}{{\sc \textbf{StoryCloze}}}
\newcommand{\storyclozel}{{\sc {StoryCloze}}}
\newcommand{\summeval}{{\sc \textbf{SummEval}}}
\newcommand{\summevall}{{\sc {SummEval}}}

\newcommand{\insted}{{\sc \textbf{INSteD}}}
\newcommand{\instedl}{{\sc {INSteD}}}
\newcommand{\wiki}{{\sc \textbf{INSteD-Wiki}}}
\newcommand{\cnn}{{\sc \textbf{INSteD-CNN}}}
\newcommand{\wikil}{{\sc {INSteD-Wiki}}}
\newcommand{\cnnl}{{\sc {INSteD-CNN}}}

\newcommand{\lmlm}{{\sc \textbf{LMvLM}}}
\newcommand{\lmlml}{{\sc {LMvLM}}}

\newcommand{\roc}{{\sc RocStories}}
\newcommand{\wsjl}{{\sc WSJ}}
\newcommand{\wsj}{{\sc \textbf{WSJ}}}

\newcommand{\unc}{{\sc UNC}}
\newcommand{\lcdg}{{\sc LCD-G}}
\newcommand{\lcdi}{{\sc LCD-I}}
\newcommand{\lcdl}{{\sc LCD-L}}
\newcommand{\lcd}{{\sc LCD}}
\newcommand{\cls}{{\textsc {[cls]}}}

\newcommand{\ie}{{\em i.e.,}\xspace}

\newcommand{\Ni}{({\em i})~}
\newcommand{\Nii}{({\em ii})~}
\newcommand{\Niii}{({\em iii})~}

\definecolor{airforceblue}{rgb}{0.0, 0.0, 0.8}
\definecolor{crimson}{rgb}{0.8, 0.0, 0.0}
\newcommand{\airforceblue}[1]{\textcolor{airforceblue}{#1}}
\newcommand{\crimson}[1]{\textcolor{crimson}{#1}}

%% file: files/math_commands.tex
\usepackage{amsmath,amsfonts,bm}

\def\eqref#1{equation~\ref{#1}}

\def\1{\bm{1}}

\def\rvq{{\mathbf{q}}}

\def\rvw{{\mathbf{w}}}

\def\rvz{{\mathbf{z}}}

\def\mQ{{\bm{Q}}}

\DeclareMathAlphabet{\mathsfit}{\encodingdefault}{\sfdefault}{m}{sl}
\SetMathAlphabet{\mathsfit}{bold}{\encodingdefault}{\sfdefault}{bx}{n}

\def\gD{{\mathcal{D}}}

\def\gL{{\mathcal{L}}}



%% file: files/introduction.tex
Coherence is a property of a well-written text that makes it different from a random set of sentences: sentences in a coherent text are connected in systematic ways such that each sentence follows naturally from  previous ones and leads into the following ones \citep{Halliday76, Grosz1986AttentionIA}. \textbf{Coherence models} \citep{Barzilay:2005} that can distinguish a coherent text from incoherent ones have a wide range of applications in language generation, summarization, and coherence assessment tasks such as essay scoring and sentence ordering.

With recent advancements in neural methods,  claims of fluency in summarization \citep{Liu2017GenerativeAN, celikyilmaz-etal-2018-deep}, language modeling \citep{GPT2-Blog, GPT-3}, response generation \citep{zhang2019dialogpt, hosseiniasl2020simple} and human parity in machine translation \citep{Hassan2018AchievingHP} have led to calls for finer-grained discourse-level evaluations \citep{Lubli2018HasMT, sharma2019entity, CUBBITT}, since traditional metrics such as BLEU and ROUGE are unable to measure text quality and readability \citep{Paulus2018ADR, reiter2018structured}. Coherence models that can evaluate machine-generated text have become the need of the hour.

A majority of coherence models proposed optimize their learning objectives on the permuted document task using the Penn Treebank (\wsjl) corpus. An original article is considered a `positive' sample of a coherent document, while a permutation of its sentences is considered a `negative' or incoherent sample (see \Cref{subsec:appendix-wsj-example} for an example). Models are usually trained in a \textit{pairwise ranking} fashion to distinguish the two.

The basic entity-grid model proposed by \citet{Barzilay:2005,Barzilay2008ModelingLC} was extended to incorporate entity-specific features \citep{Elsner:2011}, multiple ranks \citep{Feng:2012}, and coherence relations \citep{Lin:2011,Feng:2014}.  Their neural extensions have also been proposed \citep{dat-joty:2017,joty-etal-2018-coherence}. More recent state-of-the-art models like the Transferable Neural model \citep{xu-etal-2019-cross} consider coherence at a local level by training a forward and backward model only on adjacent sentences, in addition to generative pre-training of the sentence encoders. The Unified Coherence model \citep{unifiedcoherence} uses bi-linear layer and lightweight convolution-pooling in a Siamese framework to capture discourse relations and topic structures, along with an explicit language model loss to capture syntactic patterns.

\citet{rethinkingEACL} recently tested these state-of-the-art models by conducting coherence evaluations on the \wsjl\ permuted document task, machine translation, summarization and next utterance ranking tasks. They found that while models performed well on the permuted document task, when tested off-the-shelf, models generalized poorly to downstream evaluation tasks. They call for more comprehensive evaluations of coherence models. \citet{Pishdad2020HowCA} also reached a similar conclusion. They retrained several neural coherence models for tasks analogous to coherence modeling such as detecting connective substitution and topic switching. They found that {performance on the permuted document task is only partially indicative of coherence modeling capabilities}.

In light of these recent findings, our aim is to propose a coherence model that generalizes well to downstream tasks. We train our model purely through \textit{self-supervision}, without tailoring the model architecture specifically to the permuted document task {or any other form of supervision}. 

\citet{li-jurafsky:2017} point out that coherence models are exposed to a limited number of incoherent samples in the \emph{pairwise} setup, since only a small sample of all possible incoherent permutations of a document are used to train models. Learning with more negatives can better maximize the mutual information between their {representations} \citep{Oord2018RepresentationLW}. By using a \textit{contrastive learning} \citep{pmlr-v9-gutmann10a} setup, where each `positive' document is compared with multiple `negative' documents, we increase the proportion of negative samples that the model is exposed to, and show that the coherence model shows significant improvements in performance. 

\citet{Wu2020OnMI} show that the difficulty of the negative samples used for contrastive training can strongly influence model success for visual representation learning. Guided by this {principle}, we train the model with automatically mined hard negatives, coupled with a large global negative queue encoded by a momentum encoder \citep{he2019moco}.

In summary, our contributions are:
\begin{itemize}[leftmargin=*,topsep=2pt,itemsep=2pt,parsep=0pt]
        \item A neural coherence model trained purely through well-designed self-supervision tasks that generalizes well to downstream applications. 
        
        \item Evaluation on multiple independent test sets that are more indicative of real-world performance of the coherence model.
        \item Empirical results demonstrating that increase in the density and quality of negative samples leads to better generalization for coherence models.
\end{itemize}

%% file: files/datasets.tex
To ensure that our coherence model is useful for evaluation in downstream applications, we use a selection of task-independent test sets that cover a variety of domains and genres, including machine generated text from summarization systems and language models. Following \citet{Pishdad2020HowCA}, we also evaluate the models on a commonsense reasoning narrative dataset.  We train (and validate) the coherence models on standard \wsjl\ data, while using the rest as ``independent'' test sets to indicate the generalizability of the trained models. All evaluations on {downstream tasks} are conducted in a pairwise setting to enable a fair comparison.

\subsection{Training Data}
\label{subsec:data-wsj}

\paragraph{$\bullet$ \wsj} The Wall Street Journal (\wsjl{}) corpus consists of news articles divided into 1240/138/1053 documents for training/development/testing in the standard setup. We exclude documents with < 4 sentences and truncate them to a maximum length of 600 tokens. To maximally utilize documents which are otherwise truncated due to GPU memory constraints, we partition documents with 20+ sentences into blocks of 10 sentences and consider each block as a separate positive document. This increases the number of coherent `documents' that we can use to generate a larger {training} set. \citet{unifiedcoherence} use 20 permutations of a document for training; since their setup is pairwise, it means the original positive document is repeated 20x. We regenerate the permuted documents similarly, sampling a larger set of permutations for our contrastive learning setup.\footnote{We ensure that the generated permuted documents are not repeated. For example, our contrastive learning setup requires 5 negative samples per instance; because each positive document appears 20 times in the {original} dataset, 100 unique permutations would be generated and divided accordingly.} This gives us $46,522$ instances of positive and corresponding negative documents for training and $4,522$ instances for development. We use the original pairwise test set used by \citet{unifiedcoherence} with $20,411$ pairs for testing.

\subsection{Machine Generated Texts}

\paragraph{$\bullet$ \summeval}
\citet{summeval} conduct a manual coherence evaluation of the summaries generated by $16$ different summarization systems for 100 source articles based on the CNN/DailyMail \citep{Hermann2015TeachingMT} dataset. Likert-style coherence ratings from $3$ expert annotators are available for each summarized text. We adapt this to the pairwise setting by creating pairs of summaries from every system for each unique source article. The summary with the higher average coherence rating is designated as the positive document, while the summary with the lower rating is the negative document for that pair. This results in $\binom{16}{2} \times 100 = 12,000$ pairs for evaluation.

\paragraph{$\bullet$ \lmlm} 
To cover a wider variety of machine generated text, we generated texts from various language models using prompts taken from the validation and test sets of the WritingPrompts dataset \citep{WritingPrompts}. Four language models were chosen for this purpose: GPT2-Small, GPT2-XL, CTRL and GPT3. The continuations produced by these models for each prompt were truncated at approximately $150$ tokens and paired together. 
Using these texts, we conducted a user study on Amazon Mechanical Turk. Workers were instructed about the concept of coherence and shown examples of coherent and incoherent texts. Given the prompt, they were asked to choose the more coherent text out of two given language model outputs; they were also given an option to choose neither in case the texts were equally coherent/incoherent (see \Cref{subsec:appendix-user-study} for more details such as the study interface). After removing the samples with low agreements and ties, a total of $1,046$ pairs with judgments from $3$ annotators each were collected. The Krippendorff's alpha coefficient \citep{Krippendorff2011ComputingKA} between the annotators was \textbf{0.84}. We calculate the agreements of the coherence model ranking with these judgments, designated \lmlml.

\subsection{Curated Test Sets}
\label{subsec:data-curated}

\paragraph{$\bullet$ \insted} \citet{ailishen2021} propose a sentence intrusion detection task in order to test the coherence modeling capabilities of {pre-trained} language models. Incoherent documents are created by substituting a sentence from a document with another sentence from a different document, ensuring that the replacement sentence is similar to the original document to make the task sufficiently hard. We adapt their task to the pairwise setting by pairing the original coherent and the corrupted incoherent document, giving us $7,168$ instances  from their CNN test set (\cnnl{}) and $3,666$ instances from their Wikipedia test set (\wikil{}) for evaluation. \citet{ailishen2021} also create a hand-crafted linguistic probe test set, where incoherence is manually inserted based on a range of linguistic phenomena; we use this test set for analysis (\cref{sec:analysis}).

\paragraph{$\bullet$ \storycloze}
The \storyclozel\ dataset (created from \roc\ \citep{StoryCloze}) consists of a short narrative-style text with two possible endings, one of which is implausible.  The test set labels are not public so we use the validation set. We designate the text with the correct ending as the positive document and the text with the incorrect ending as the negative document, resulting in a total of $1,571$ pairs for evaluation.

%% file: files/experiments.tex
\subsection{Model Architecture}

Previous work on coherence modeling proposed elaborate architectures to capture various aspects of coherence (see \cref{sec:intro}). {However, our {key} hypothesis is that large-scale pre-trained models are expressive enough to model coherence given the right self-supervision.} Effective bi-directional encoding through large Transformer networks \citep{VaswaniNIPS2017} can consider longer language context, while language modeling objectives enforce syntactic and local coherence patterns in the model.

In our work, we adopt XLNet \citep{XLNet} as the backbone model. It is trained using a permuted language modeling objective, in which the expected log-likelihood of a sequence with respect to all permutations of the factorization order is maximized. This allows the modeling of bi-directional context, while maintaining the auto-regressive property and avoiding the pretrain-finetune discrepancy. In addition, XLNet also incorporates segment recurrence (or memory) and the relative encoding scheme of Transformer-XL \citep{Dai2019TransformerXLAL}, which makes it effective in modeling longer text sequences. This makes it suitable for our purpose of coherence modeling. 

Given a document $\gD$ with $n$ sentences $(s_1, s_2, \dots, s_n)$ as input, our model uses the representations obtained through XLNet (parameterized by $\phi$) to assign a coherence score to the model. Specifically, for each sentence $s_i$ with $k$ tokens $(w_1, w_2 \dots w_k)$, XLNet maps {each token $w_t$} to its vector representation $v_t \in \mathbb{R}^d$ where $d$ is the dimension of the embedding. In addition, the complete input $\gD$ is also mapped to a document representation $\rvz  \in \mathbb{R}^d$ (\ie\ the representation of the $\cls$ token). We simply add a linear layer to convert document representation $\rvz$ to obtain the final coherence score: $f_{\theta}(\gD) = \mathbf{w}^\top \rvz + b$, where $\mathbf{w}$ and $b$ are the weight and bias of the linear layer with {$\theta  = \{\phi, \rvw, b \}$} being the entire parameter set of the model (see the upper part of \Cref{fig:coherence-moco}).

\subsection{Margin-based Pairwise Ranking}
\label{subsec:pairwise}
\paragraph{Setup.} 
Traditionally, coherence model training has been done in a pairwise ranking setup. In this setup, the model is trained to score the coherent or positive document higher than the incoherent or negative document, using a pairwise ranking loss \citep{collobert2011natural} defined as follows:
\begin{equation}
\small
\gL_{\theta} = \max \big(0, \tau - f_{\theta}(\gD^{+}) + f_{\theta}(\gD^{-}) \big)
\normalsize
\end{equation}
where $f_{\theta}(\gD^{+})$ is the coherence score of the positive document, $f_{\theta}(\gD^{-})$ is the coherence score of the negative document and $\tau$ is the margin.

\paragraph{Baselines.} We compare our models against all three versions of the \textbf{L}ocal \textbf{C}oherence \textbf{D}iscriminator or LCD model \citep{xu-etal-2019-cross}\footnote{{\href{https://github.com/BorealisAI/cross_domain_coherence}{https://github.com/BorealisAI/cross\_domain\_coherence}} }: \Ni \lcdg{}, that uses GloVe \citep{pennington2014glove} representations, \Nii \lcdi{}, that uses InferSent \citep{Conneau2017SupervisedLO} representations, and \Niii \lcdl{}, that uses representations from an RNN-based language model trained on the training data. We also compare against the \textbf{Un}ified \textbf{C}oherence model or \unc{} \citep{unifiedcoherence}\footnote{{\href{https://github.com/taasnim/unified-coherence-model}{https://github.com/taasnim/unified-coherence-model}}}, which is the previous SOTA on the WSJ permuted document task. {Results from evaluation of existing coherence models by \citet{Pishdad2020HowCA} and \citet{rethinkingEACL} indicate that \unc{} and LCD are the best-performing models {(see \Cref{subsec:appendix-prev-evals} for a full comparison).}} We retrain their models with our training data for comparison. In addition, to ascertain the contribution of the pre-trained XLNet embeddings, we train our pairwise model without fine-tuning the representations, \ie\ only the score-producing linear layer weights $\mathbf{w}$ and $b$ are trained {on the pairwise ranking task}.

\paragraph{Results.} The results for the baseline models  are given in \Cref{tab:main-results} (see top five rows). {We see that despite accuracies of more than 90\%  on the \wsjl\ permuted document task, the \lcd\ models perform only a little above a random baseline of 50\% on most of the independent test sets, with \lcdg{} being the best generalizing model out of the three.  Similarly, despite a relatively high performance on the \wsjl\ test set (94.11\%), {\unc's} performance on the independent test sets is quite poor, even failing to do better than the random baseline of 50\% in {two out of five cases}. Both the \lcd\ and \unc\ models have slightly better success on the \cnnl\ dataset, which is the same domain (news) as the training data, with the \unc\ model reaching 67.21\% accuracy. Our XLNet-Pairwise model trained without fine-tuning the representations {(No FT)} performs no better than the baseline models. This shows that both the \lcdg\ and the \unc\ models are in fact strong baselines despite using GloVe and ELMo \citep{Peters:2018} pretrained representations respectively. }

{Our fully-trained XLNet-Pairwise model } not only outperforms the \unc{} model on the {standard} \wsjl{} permuted document task, but also significantly outperforms all baseline models on the independent test sets, showing an absolute improvement of 15-20\% on the \summevall{}, \cnnl{}, \wikil{} and the \storyclozel\ datasets.  On \lmlml{}, the \unc{} model has a better performance; we suspect that its explicit conditional language modeling loss might provide an additional advantage for this particular task. Overall, our results are consistent with observations from \citet{rethinkingEACL} that show poor generalizability in the previous SOTA model.

\begin{table*}[h]
    \centering
    \scalebox{0.8}{\begin{tabular}{l|c|c|c|c|c|c}
    \toprule
    
      \textbf{Model }& \textbf{\wsj} & \summeval & \lmlm & \cnn & \wiki & \storycloze \\
      
      \midrule

      \lcdg\ & $90.39_{\pm 0.28}$ &	$54.15_{\pm 0.83}$ &	$0.419_{\pm 0.00}$ &	$61.24_{\pm 0.71}$ &	$55.09_{\pm 0.46}$  & $51.76_{\pm 1.22}$ \\
      
      \lcdi\ & $91.56_{\pm 0.16}$ &	$51.71_{\pm 0.99}$ &	$0.420_{\pm 0.01}$ &	$60.23_{\pm 0.86}$ &	$53.50_{\pm 0.37}$ &	$52.69_{\pm 0.69}$ \\
      
      \lcdl\ & $90.24_{\pm 0.36}$ &	$53.56_{\pm 1.20}$ &	$0.404_{\pm 0.01}$	& $55.07_{\pm 0.26}$ &	$51.04_{\pm 0.47}$ & $50.09_{\pm 1.57}$\\
      
      \unc\  & $94.11_{\pm 0.29}$	&$46.28_{\pm 0.80}$		&$0.463_{\pm 0.01}$	&$67.21_{\pm 0.55}$	&$55.97_{\pm 0.45}$ &$49.39_{\pm 1.81}$ \\

      \rowcolor{LRed} Our - Pairwise (No FT) & $71.70_{\pm 1.02}$ &	$54.93_{\pm 1.91}$ &	$0.421_{\pm 0.01}$ &	$59.96_{\pm 3.15}$ &	$53.45_{\pm 0.86}$ &	$51.69_{\pm 1.32}$  \\ 
     
       Our - Pairwise & $98.23_{\pm 0.20}$ & $64.83_{\pm 1.03}$  &	$0.458_{\pm 0.02}$ & $91.96_{\pm 1.09}$ &	$70.85_{\pm 1.85}$  &	$71.84_{\pm 2.33}$\\

\rowcolor{LGreen} Our - Contrastive & $98.59_{\pm 0.20}$	& $66.93_{\pm 1.10}$	& $0.468_{\pm 0.01}$ &	$92.84_{\pm 0.61}$ &	$71.86_{\pm 0.69}$ 	& $72.83_{\pm 2.89}$ \\
    
\rowcolor{LBlue} Our - Full Model & $98.58_{\pm 0.18}$ & $67.19_{\pm 0.63}$ &	$0.473_{\pm 0.00}$ &	$93.36_{\pm 0.49}$ &	$72.04_{\pm 1.05}$ &	$74.62_{\pm 2.79}$ \\
\bottomrule
\end{tabular}}

\caption{Results on the \wsjl\ permuted document test set and the various independent test sets of {\lcd\ GloVe (LCD-G), \lcd\ {I}nfersent (LCD-I), \lcd\ Language Model (LCD-L)}, \unc{}, and our XLNet based models. {The XLNet representations are not fine-tuned during training for our Pairwise (No FT) model.} {Except for the \lmlml{} results which are reported in terms of Krippendorff's alpha agreement with human annotators, all other results are reported in terms of accuracy of the models in scoring the positive document higher than the negative document.} All results are averaged over 5 runs with different seeds.} 
\label{tab:main-results}

\end{table*}

\subsection{Contrastive Learning}
\label{subsec:contrastive}

\paragraph{Setup.} 
In pairwise ranking, each positive sample is only compared to one negative at a time. Contrastive learning \citep{pmlr-v9-gutmann10a} makes it general, where a single positive sample can be compared to multiple negatives, which can be particularly useful in the permuted document task where the number of possible incoherent samples per coherent document can be very large. The number of negatives considered and their quality can affect model performance \citep{pmlr-v97-saunshi19a}. \citet{Wu2020OnMI} show that contrastive loss maximizes a lower bound on the mutual information between representations. A larger number of negatives increases the tightness of the bound; learning with more negatives can better maximise the mutual information. We train our model with a margin-based contrastive loss defined as:
\begin{align}
\label{eqn:contrastive}
\small
\begin{split}
\gL_{\theta} = -\log\Big(\frac{e^{f_{\theta}(\gD^{+})}}{e^{f_{\theta}(\gD^{+})} + \sum_{j=1}^N e^{(f_{\theta}(\gD^{-}_{j})-\tau)}}\Big)
\end{split}
\end{align}
\noindent where $f_{\theta}(\gD^{+})$ is the coherence score of the positive document, $f_{\theta}(\gD^{-}_{1}),\cdots, f_{\theta}(\gD^{-}_{N})$ are the scores of the $N$ negative documents, and $\tau$ is the margin.

\paragraph{Training.} We use the same training data as the baseline models to train our contrastive model; the positive documents remain the same, while we use 5 negative documents per instance (instead of only 1 in the pairwise setup). Effectively, the model sees the same number of positive or coherent documents, but five times as many negative samples during training compared to the pairwise setting.  \Cref{subsec:appendix-hyperparams} gives the full set of hyperparameters.

\paragraph{Results.} From the results in \Cref{tab:main-results}, we see that the contrastive model {(second to last row)} further improves the results across all the independent test sets; the results on the \lmlml{} dataset also improve, surpassing the \unc\ model performance. Although the improvement on the \wsjl\ permuted document task is small, the improvement in the generalizability of the model is more significant.

\subsection{{Hard Negative Mining}}
\label{subsec:hardneg_mining}

It has been shown that the difficulty of the negative samples used for contrastive training can strongly influence model success {\citep{Wu2020OnMI,Huang20}}. We therefore automatically mine hard negatives during training. For the permuted document task, we can take advantage of the fact that the negative sample space can be huge; for a document with $n$ sentences, the candidate pool of permutations has $n!-1$ incoherent documents from which we can mine hard negatives. {For the problem of dense text retrieval,} \citet{Xiong2021ApproximateNN} find \textit{global} hard negatives by computing document encodings using a recent checkpoint to build an asynchronous index of the entire corpus, and sampling negative documents from the index.  However, the huge candidate pool for permuted documents also makes it infeasible to mine global negatives in our case. 

Instead, we perform \textit{local} negative sample ranking. For each positive instance in the training data, we sample a larger number of permuted documents ($h$) per instance than we need for training {(\ie\ $h > N$)}. We score these negative documents using the model updated thus far and use the highest ranking negatives for training. Specifically, the model is first trained with $x$ instances ($x$ is a hyperparameter) of data, by using 5 negative samples randomly chosen out of $h$. The updated model is then used to score all the $h$ negative samples each for another set of $x$ instances from the training data. The scores of the $h$ negative samples are ranked and the top scoring 5 negative samples for each instance are used to train the model for the next $x$ {gradient} steps. This process is repeated throughout training; the model therefore iteratively mines harder and harder negative samples as it improves. See \Cref{alg:hard-neg} in \Cref{subsec:appendix-pseudocode} for the pseudocode.

{In practice however, we find that using hard negative samples directly leads to instability in model training (see \Cref{subsec:analysis-hardneg}). We therefore use hard negative training in combination with a momentum encoder, which we describe in the next subsection.}

\begin{figure*}[t!]
    \centering
    \includegraphics[scale=0.47]{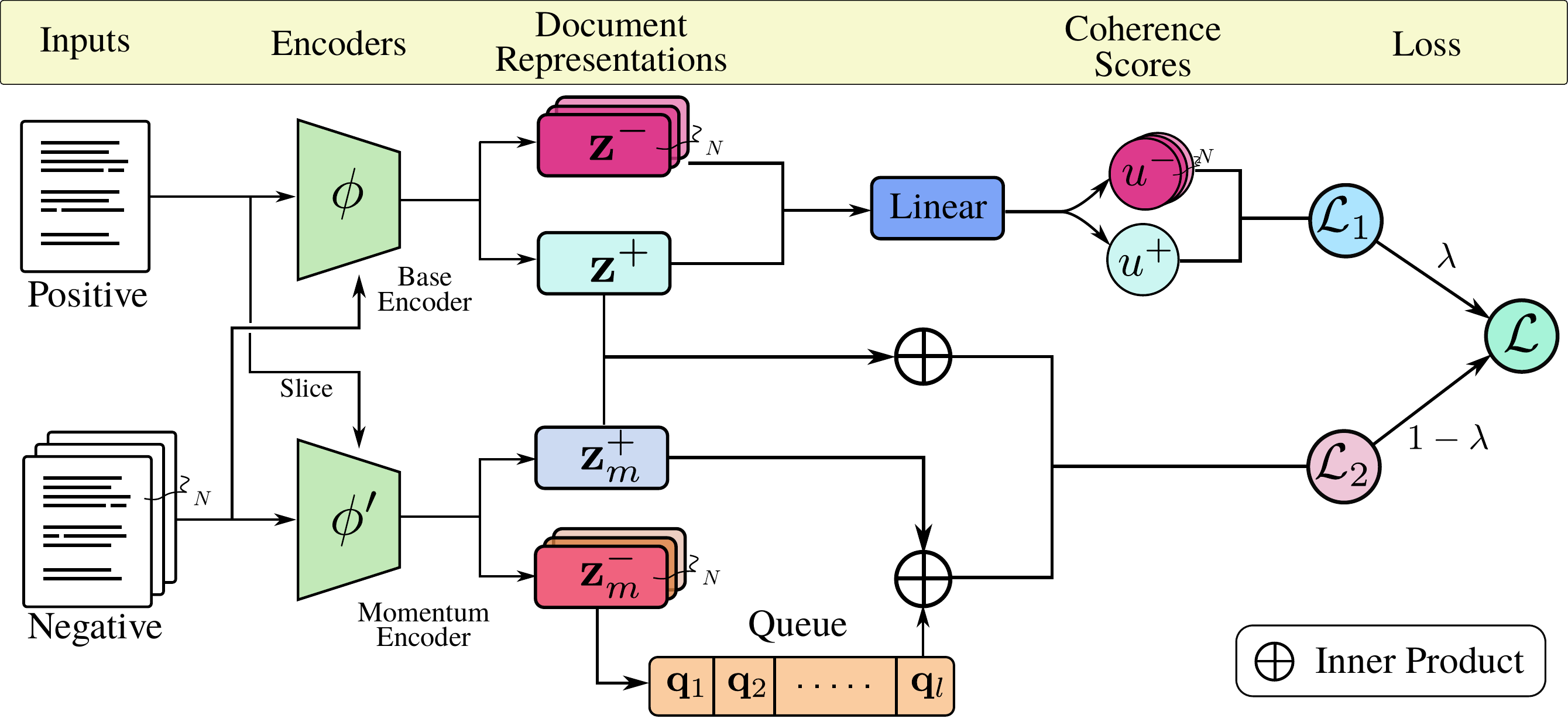}
       \caption{Our coherence model with the auxiliary momentum encoder. $\phi$ is our base encoder similar to our setup in \Cref{subsec:contrastive}, while $\phi'$ is our momentum encoder. $u^{+} = f_\theta (\gD^{+})$ and $u^{-} = f_\theta (\gD^{-})$ are the coherence scores of the positive and negative documents respectively. Note that only the parameters of $\phi$ and the linear layer are updated through backpropagation.}
    \label{fig:coherence-moco}

\end{figure*}

\subsection{{Hard Negatives with Momentum Encoder}}
\label{subsec:momentum_model}

While increasing the number of negative samples per instance has been shown to be effective for constrastive learning, resource constraints can limit the number of negatives that can be considered per instance. One solution is to consider other positive instances in the same training batch as negatives \citep{Karpukhin2020DensePR, Chen2020ASF}. However, it is not suitable for the permuted document task since the negatives are instance-specific. While a permuted document is still independently incoherent, training with permutations of other documents will not provide the same cues for coherence modeling as the original self-supervision.

Another solution is to maintain a large global queue of negative samples that are independent of the current training instance. During training, negative samples {(specifically, their representations)} from the latest batch are enqueued to build a queue upto some size $l$. As training continues, the negative samples from the oldest batch are dequeued to accommodate newer samples. However, representations of the documents will evolve through training {as the model parameters get updated}; this will make the negative samples in the queue inconsistent with each other and the training instances {in the current batch}. Moreover, the issue of mismatched self-supervision with negatives that are permuted versions of other documents still remains.

\paragraph{Momentum Encoder.} To address these issues, we add an auxiliary momentum encoder \citep{he2019moco}, which is also XLNet \citep{XLNet}. {\Cref{fig:coherence-moco} shows the overall architecture.} Keeping the base contrastive setup the same (the upper part), we add an additional contrastive objective based on representations from the momentum encoder. Specifically, we re-encode the positive and negative samples through the momentum encoder; the negative samples thus encoded are used to build the queue. We train the model to promote the similarity between the positive representations from the momentum encoder and the positive representations from our base encoder over the similarity with the negative samples from the queue, $\mQ$. {Specifically, we define a momentum loss $\gL_{\theta}^{\text{mom}}$ as:}
\begin{align}
\small 
\begin{split}
 c^{+} = \frac{(\rvz^{+})^\top (\rvz^{+}_{m})}{||\rvz^{+}||  ~||\rvz^{+}_m||}; \hspace{1em}
 c^{-}_j = \frac{ (\rvz^+_m)^\top  \rvq_j}{||\rvz^{+}_m||  ~||\rvq_j||};\\
 \gL_{\theta}^{\text{mom}} =  -\log\Big(\frac{e^{c^{+}}}{e^{c^{+}} + \sum_{j=1}^l e^{(c^{-}_j-\tau)}}\Big)    \label{eq:moco}
\end{split}
\end{align}
where $\rvz^{+}$ and $\rvz^{+}_m$ are the positive representations from the base encoder ($\phi$) and the momentum encoder ($\phi'$) respectively, $\rvq_1, \dots, \rvq_l$ indexed by $j$ are the negative representations from $\phi'$ in the queue, and $\tau$ is the margin. 
The momentum encoder $\phi'$ is updated based on the base encoder $\phi$ as:
\begin{equation}
    \phi' \leftarrow \mu * \phi' + (1 - \mu) * \phi
\end{equation}
where $\mu \in [0,1)$ is the momentum coefficient; only $\phi$ is updated through backpropagation. Our full model is trained with a combination of the original contrastive objective (Eq. \ref{eqn:contrastive}) and the momentum encoded contrastive similarity objective (Eq. \ref{eq:moco}):
\begin{equation}
\label{eq:combined_loss}
\gL_{\theta} = \lambda \gL_{\theta} + (1-\lambda) \gL_{\theta}^{\text{mom}}
\end{equation}
where $\lambda$ is a weighting hyperparameter. {Note that the momentum encoder can be considered as a \emph{temporal ensemble} model consisting of exponential-moving-average versions of the base model. Due to this, the gradients from the momentum loss (Eq. \ref{eq:moco}) also help in stabilising the overall training (\Cref{sec:analysis}).}

\paragraph{Length Invariance.} In the permuted document task, both the positive and the negative samples have the same number of sentences. This is not necessarily the case for downstream applications. To incorporate length invariance into our model, we encode a random contiguous slice of the positive document through the momentum encoder $\phi'$.\footnote{Minimum is 4 and maximum is full document.}

The global negatives queue $\mQ$ is constructed from the mined hard negative samples used for training. Our model is therefore trained to rely not only on comparative coherence cues from the traditional permuted document setup, but also to recognize more independent cues for coherence through the global queue, which is additionally enhanced by incorporating length invariance and automatically mined hard negative samples.

\paragraph{Training.} We train the model with the same training data, this time sampling $h=50$ negatives\footnote{As previously described in \Cref{data}, we ensure the sampled negative documents are unique even when the positive documents are repeated. This ensures that a much larger sample of the overall candidate pool is considered during training. Since we sample and rank 50 negative documents per positive instance, accounting for 20 repetitions of the positive documents, $20 * 50 = 1000$ total negative documents are considered for hard negative mining. This is 10 times larger than the contrastive setup (100 unique negatives) and 50 times larger than the pairwise setup (only 20 unique negatives). } per instance for hard negative ranking, and setting the training steps {(or instances)} $x=200$. We use a queue size of $l=1000$ and set our momentum coefficient $\mu=0.9999999$, with loss weighting parameter $\lambda=0.85$. {Due to GPU memory constraints (24GB, Quadro RTX 6000), we train our model with a batch size of 1.} See \Cref{subsec:appendix-hyperparams} for the full set of hyperparameters.

\paragraph{Results.} The results in \Cref{tab:main-results} (last row) show that our momentum encoder model with hard negative mining outperforms all previous models across the independent testsets. This improvement comes despite a very similar performance on the \wsjl\ test set; we believe that our model truly improves in generalizability without overfitting to the permuted document task. The improvements on the out-of-domain test sets, particularly on \lmlml\ and \storyclozel{}, support this conclusion.

%% file: files/analysis.tex
\begin{figure*}[t]
    \centering
    \begin{subfigure}{0.50\linewidth}
    \centering
    \includegraphics[width=\linewidth]{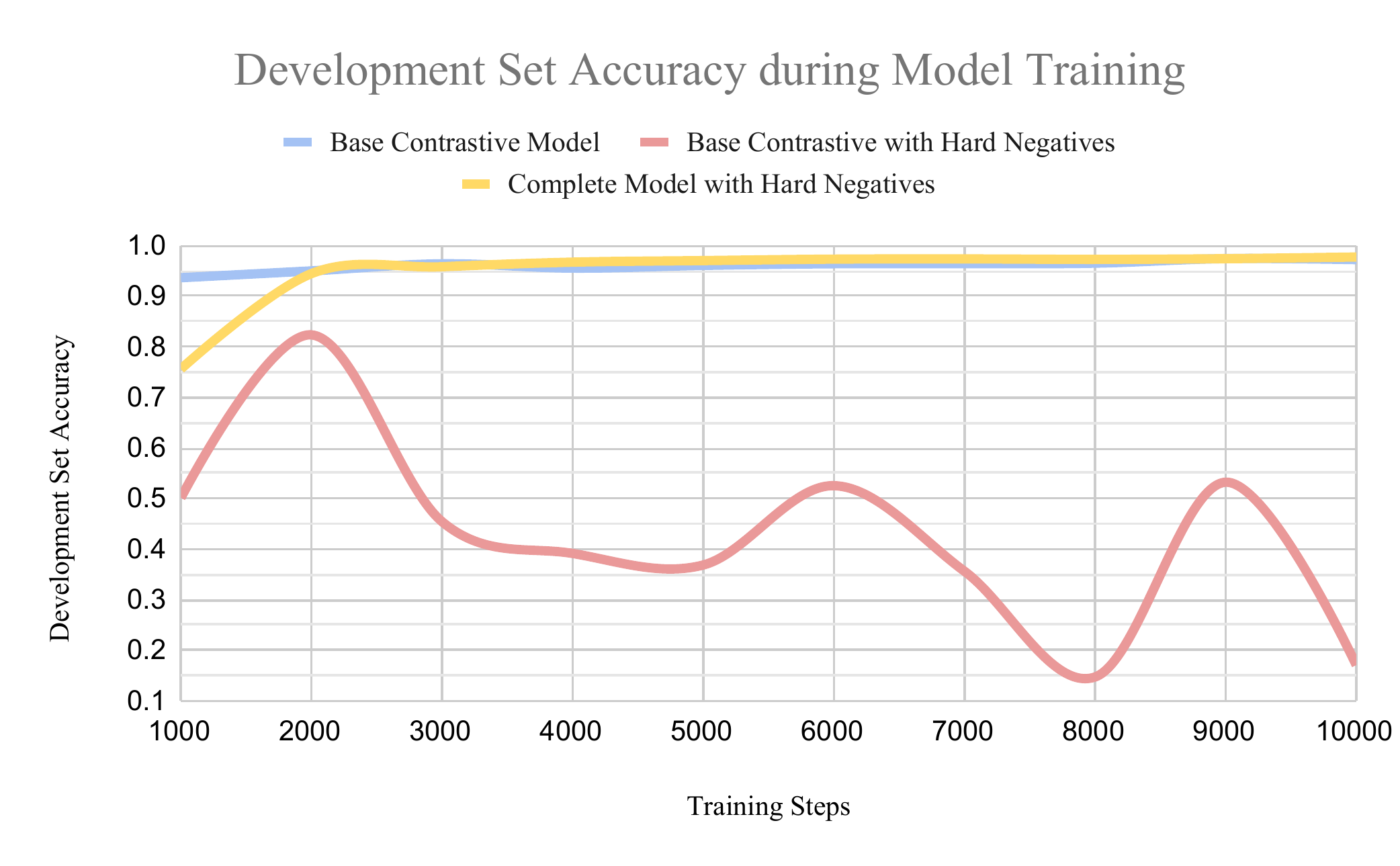}
    \caption{}
    \label{fig:moco_vs_hardneg}
    \end{subfigure}%
     \begin{subfigure}{0.46\linewidth}
     \centering
     \includegraphics[width=\linewidth]{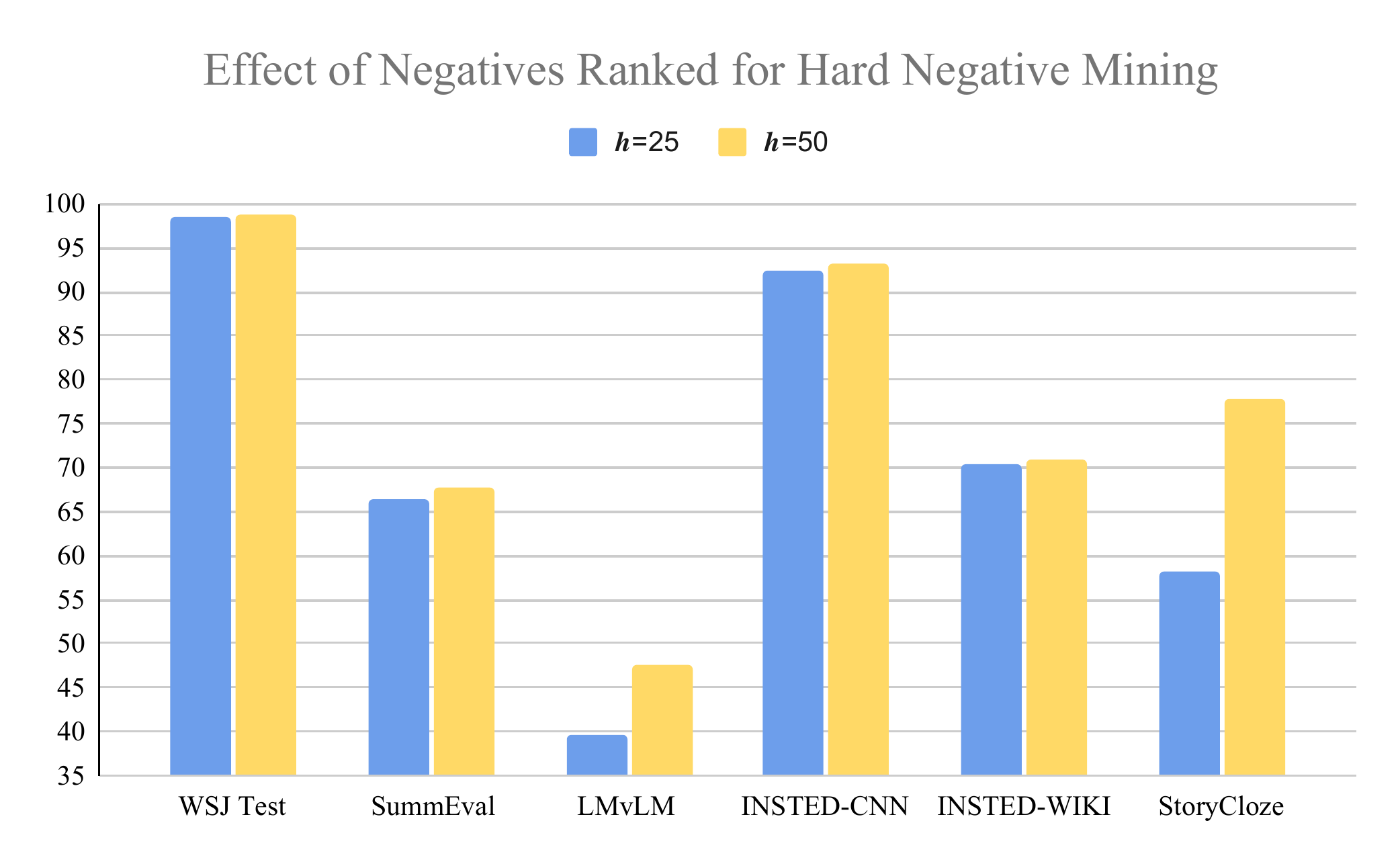}
     \caption{}
    \label{fig:rank_neg}
    \end{subfigure}
    \begin{subfigure}{0.48\linewidth}
    \centering
    \includegraphics[width=\linewidth]{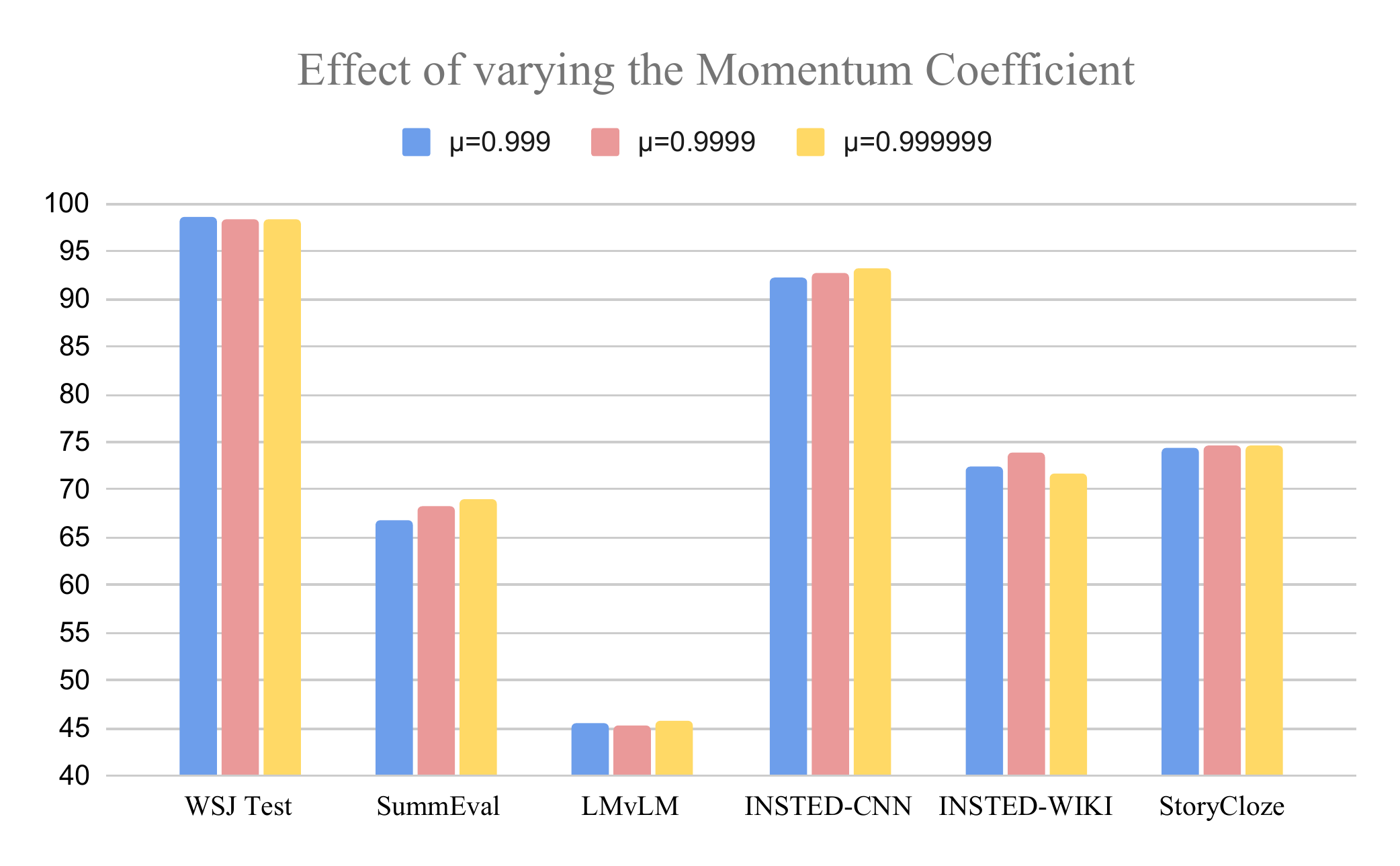}
    \caption{}
     \label{fig:mocoefficient}
    \end{subfigure}%
    \begin{subfigure}{0.48\linewidth}
    \centering
     \includegraphics[width=\linewidth]{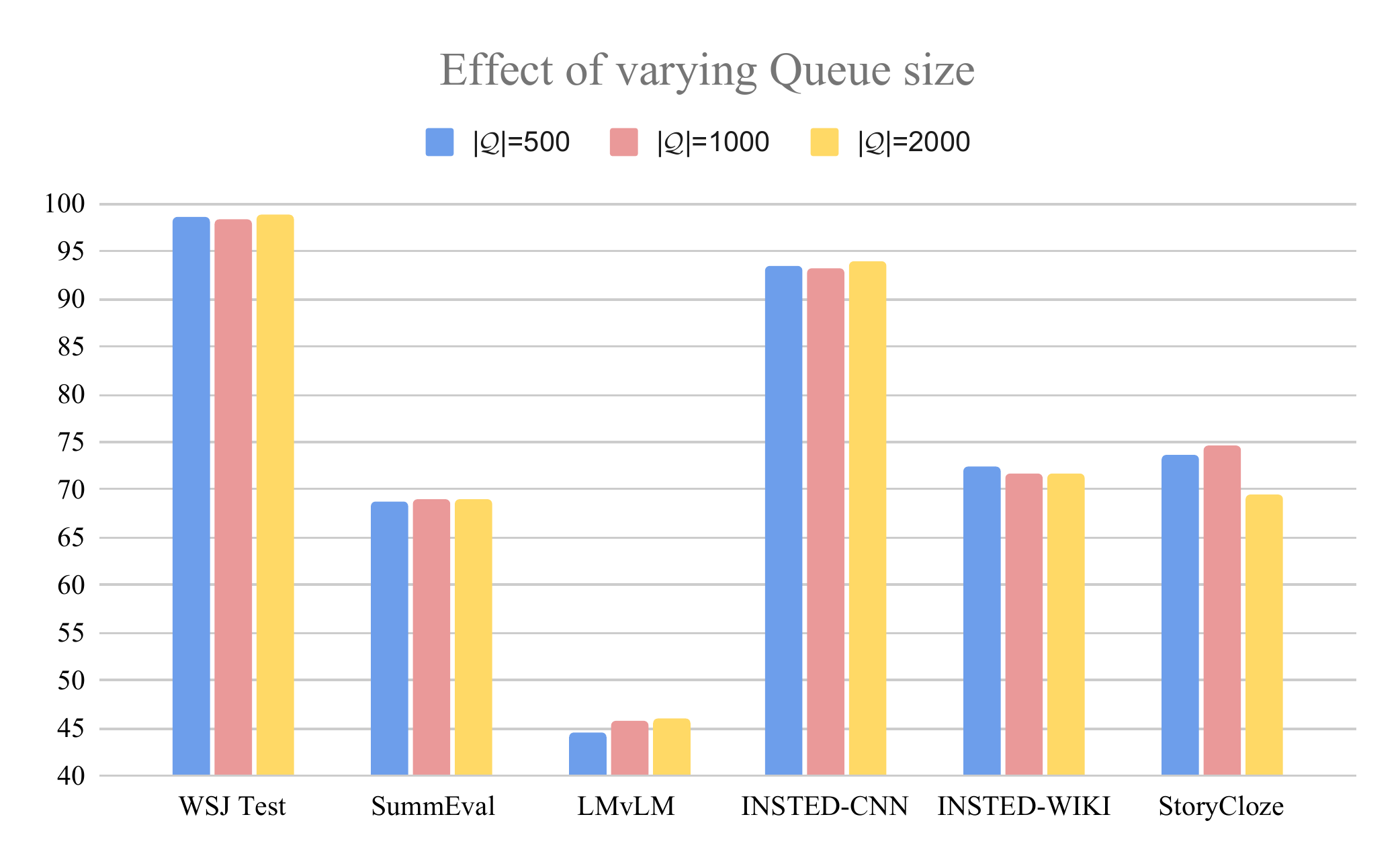}
     \caption{}
    \label{fig:queue_plot}
    \end{subfigure}
    \caption{(\subref{fig:moco_vs_hardneg}) A plot of the development accuracy during training our contrastive model with and without hard negative mining, and our complete model with hard negative mining. The accuracies are evaluated after every 1000 {gradient} steps. (\subref{fig:rank_neg}) Results on the various test sets for our model trained with hard negative mining by sampling different number of negatives ($h$) for ranking.  (\subref{fig:mocoefficient}) Results on the various test sets for our complete model trained with different momentum coefficient ($\mu$) values. (\subref{fig:queue_plot}) Results on the various test sets for our model trained with different global queue $\mQ$ sizes. Please note that the agreement values for \lmlml\ test set have been scaled by a factor of 100 to facilitate visualization in figures (\subref{fig:rank_neg}), (\subref{fig:mocoefficient}) and  (\subref{fig:queue_plot}).}
    \label{fig:my_label}
 
\end{figure*}

\subsection{Hard Negative Training}
\label{subsec:analysis-hardneg}

We only train our complete model (\ie\ base contrastive plus momentum model) by mining hard negative samples (\Cref{subsec:momentum_model}), because we find that training the base contrastive model directly with hard negatives leads to instability during training. \Cref{fig:moco_vs_hardneg} plots development set accuracies of our base model trained with and without hard negative mining, and our complete model trained with hard negative mining (evaluated every 1000 steps). As seen in the figure, the contrastive model displays significant volatility when trained with hard negatives {only}, while the complete model is quite stable. {This is inline with the finding of \citet{Xuan2020HardNE} who show that training with the hardest negative samples leads to bad local minima. This can be explained with the gradient analysis of such negatives which have a larger gradient norm \cite{Xiong2021ApproximateNN}, resulting in abrupt gradient steps. The momentum encoder being a temporal ensemble of the base models has a regularizing effect, addressing this issue and leading to stable and improved results (see \Cref{subsec:momentum_model}).}

\begin{table*}[th]
    \centering
    \scalebox{0.71}{\begin{tabular}{c|c|c|c|c|c|c|c|c}
    \toprule
    
      \textbf{Train Dataset}& \textbf{Neg. Type} &\textbf{Model} &\textbf{\wsj} & \summeval & \lmlm & \cnn & \wiki &\storycloze \\
      \midrule
      
      \rowcolor{LGreen} \wikil & Intrusion & Pairwise &  $95.24_{\pm 0.37}$ &	$53.03_{\pm 1.49}$ &	$0.490_{\pm 0.01}$ &	$94.07_{\pm 0.29}$ &	$82.01_{\pm 0.24}$ &	$64.21_{\pm 1.98}$\\
      
      \rowcolor{LGreen} \cnnl & Intrusion & Pairwise & $95.48_{\pm 0.47 }$ &	$57.85_{\pm 2.47}$ &	$0.502_{\pm 0.01}$ &	$97.83_{\pm 0.15}$ &	$73.52_{\pm 1.17}$ &	$71.75_{\pm 1.81}$  \\
      \midrule
      
      \rowcolor{LBlue} \wikil & Permuted & Pairwise & $96.89_{\pm 0.23}$ &	$64.53_{\pm 0.82}$ &	$0.491_{\pm 0.01}$ & $84.17_{\pm 1.50}$ &	$71.35_{\pm 0.88}$ &	$69.09_{\pm 2.29}$ \\

      \rowcolor{LBlue} \cnnl\ & Permuted & Pairwise & $97.03_{\pm 0.12 }$	&$66.63_{\pm 0.97}$	&$0.483_{\pm 0.01}$	&$92.61_{\pm 0.62}$	&$69.88_{\pm 0.64}$	&$68.95_{\pm 1.02}$ \\

     \rowcolor{LBlue} \wsjl\ & Permuted & Pairwise & $98.23_{\pm 0.20}$ & $64.83_{\pm 1.03}$  &	$0.458_{\pm 0.02}$ & $91.96_{\pm 1.09}$ &	$70.85_{\pm 1.85}$  &	$71.84_{\pm 2.33}$\\

      \bottomrule
      
\end{tabular}}
\caption{Results on the \wsjl\ permuted document test set and other independent test sets of the pairwise model trained on different datasets. All results are averaged over 5 runs with different seeds.} 
\label{tab:diff-task-results}
\end{table*}

\begin{table*}[t]
    \centering
    \scalebox{0.67}{\begin{tabular}{l|c|c|c|cl}
    \toprule
        \textbf{Linguistic Probe} & \textbf{\lcd{}} &\textbf{\unc{}} & \textbf{Our}&  & \textbf{Example} \\
    \midrule
       Pronoun Animacy Downgrade & 87.0 & 76.0  & 100.0 & \cmark & \airforceblue{She}$\rightarrow$\crimson{It} was the mother of twins Lakshmana and Shatrughna.\\
       Pronoun Animacy Upgrade & 46.0 & 63.0 & 100.0 & \cmark & \airforceblue{It}$\rightarrow$\crimson{She} has been collected in two tankōbon volumes.\\
       Pronoun Gender Flip & 49.0 &55.0 & 100.0 & \cmark & \airforceblue{She}$\rightarrow$\crimson{He} is also well known for \airforceblue{her}$\rightarrow$\crimson{his} role as Mary, the mother of Jesus.\\
       Past to Future Flip & 68.0 & 86.0 & 96.0 & \xmark & The Danes \airforceblue{finished}$\rightarrow$\crimson{will finish} first in the 2014 World Junior Hockey Championship.\\
       Single Determiner Flip &57.9 & 62.1 & 83.2 & \xmark & In 1969, he was again sold, \airforceblue{this}$\rightarrow$\crimson{these} time to the Milwaukee Bucks.\\
       Number & 56.0 & 58.0 & 80.0 & \xmark &  He had a career record of \airforceblue{67}$\rightarrow$\crimson{6.7} wins and \airforceblue{62}$\rightarrow$\crimson{-6.2} losses.\\
       Conjunction Flip & 54.0 & 55.0 & 78.0 & \xmark & The school was founded in 1908, {\airforceblue{and}$\rightarrow$\crimson{but}} has been a non-profit organization since 1956. \\
       Negation & 46.0 &60.0 & 78.0 & \xmark & He was \crimson{not} named as the Australian squad captain and was \crimson{not} captain of the Wallabies.\\

       \bottomrule

    \end{tabular}}
\caption{{Accuracies of the best performing \lcdg{}, \unc\ and our full model} on the hand-crafted linguistic probe dataset constructed by \citet{ailishen2021}. Examples (abridged for brevity) shown indicate the manual changes made to make the text incoherent; the original words are shown in \airforceblue{blue} while the modified/added words are shown in \crimson{red}. Checks (\cmark) indicate our model correctly scored the coherent text higher for that example, while crosses (\xmark) indicate that our model failed to do so.}
    \label{tab:ling_probe}

\end{table*}

\subsection{Effects of Hyperparameters}

\paragraph{Number of Ranked Negatives.} \Cref{fig:rank_neg} shows the results across the test sets for different numbers of negative samples considered for ranking ($h$) during hard negative mining. We see that increasing the number of negatives considered improves results across the board, with results on out-of-domain test sets \lmlml\ and \storyclozel\ showing particular improvement.

\paragraph{Momentum Coefficient.} \Cref{fig:mocoefficient} shows the variation in the model performance across the test sets for different values of the momentum coefficient $\mu$. We see that apart from a slight drop on the \wikil\ dataset at $\mu=0.9999999$, overall an increasing $\mu$ value leads to better generalization on the independent test sets, presumably due to a more consistent global negative queue.

\paragraph{Queue Size.}
\Cref{fig:queue_plot} shows the variation in model performance across different test sets for various sizes of the global negative queue $\mQ$. We see that while increasing the queue size generally leads to an improvement in scores, at high queue sizes the improvement is limited to test sets from the same domain (\wsjl{}, \summevall{} and \cnnl{}), and the model's generalizability is affected. 

\subsection{Effects of Varying Task \& Dataset}

\label{subsec:vary_data}

So far, we have reported the results of training our model on the permuted document task using documents from the \wsjl\ corpus as was done by most prior work \citep{Elsner:2011,unifiedcoherence}. We now test the effectiveness of other datasets, by varying the task itself and by using a different dataset for the permuted document task.

\paragraph{Sentence Intrusion.} As described in \Cref{subsec:data-curated}, \citet{ailishen2021} propose a sentence intrusion task to test coherence modeling capabilities of pre-trained language models. 
We adapt their dataset to the pairwise setting {by pairing the original coherent document (positive) with the corrupted (negative) document; setting aside 10\% of the data for development gives us 25,852 positive-negative training pairs for \cnnl\ and 41,135 pairs for \wikil{}.} We train our {pairwise (\Cref{subsec:pairwise}) model} on this task. From the results in \Cref{tab:diff-task-results} {(first two rows)}, we see that the performance on the same domain/task (as the training) and the performance on the \lmlml\ dataset is high, but the models trained on this task generalize poorly to the other independent test sets. 

\paragraph{Permuted Document Task with \insted{}.} We train our model on the permuted document task using the \instedl\ datasets. {We generate 52,607 and 66,679 positive-negative pairs for \cnnl\ and \wikil\ respectively by sampling permutations, similar to our training data (see \Cref{subsec:data-wsj}), and train our pairwise model with this data. } Specifically for machine generated texts, results in \Cref{tab:diff-task-results} show that the sentence intrusion task training does better on the \lmlml\ dataset. On the other hand, the permuted document task training does better on \summevall{}. This could be because the documents in \summevall{} are summaries of the same source article and therefore similar in content (detecting incoherence through permutations might help here), while the text generated by language models even for the same prompt tends to differ in content more significantly (detecting intruder sentences might help here). 
{Additionally, the performance of our \wsjl\ model on the \cnnl\ and \wikil\ datasets is comparable to the performance of the respective in-domain pairwise models, while outperforming both the other models on the \storyclozel\ dataset. Overall, the  model trained on the \wsjl\ permuted document task generalizes well. }

\subsection{Linguistic Probe Analysis}

\citet{ailishen2021} create 8 hand-crafted linguistic probe test sets by manually modifying words in coherent texts based on various linguistic phenomena, ensuring the incoherent text produced as a result remains syntactically correct. Except for the words targeted by the probe, the rest of the text remains identical. Each test set has 100 samples each.\footnote{Except for Single Determiner Flip, which has 95.}

We evaluate {the best performing \lcdg{}, \unc\ and our full models}  on these test sets. The results are shown in \Cref{tab:ling_probe} along with some examples from the dataset. {The \lcdg\ model has mixed success across the test sets. The \unc\ model  has the most success with the tense agreement test set and is moderately successful on the pronoun test sets.} We see that our model has perfect accuracy on all pronoun-related test sets and near-perfect accuracy on the tense agreement test set. This shows that our model is indeed capturing the discourse-level phenomena that constitute coherence. Where our model falters is in cases which may require commonsense knowledge, such as identifying that \textit{6.7 wins} is not possible. Overall, our model is quite successful in detecting several kinds of incoherence.

%% file: files/conclusions.tex
We show empirically that increasing the ratio and quality of negative samples improves the generalizability of the coherence model. We also test our model on a wide-ranging collection of independent test sets that resemble {downstream} applications, including machine generated text, on which our model significantly outperforms the previous SOTA model. Our work thus also sets a new evaluation standard for future research in coherence modeling. {We open source our code base to encourage research in a new paradigm of coherence modeling.}

%% file: files/ethics.tex
\subsection*{Data}

A description of the data pre-processing is provided in \Cref{subsec:data-wsj}. Datasets that we created will be open-sourced. In the case of the \wsjl\ dataset, the data is licensed for use only to members by the Linguistic Data Consortium. Consequently, we only release scripts to generate the data we use and not the data itself. We highlight however that the permuted document self-supervision task that we train on is independent of the dataset used and the task can be reproduced on any other corpus; see also \Cref{subsec:vary_data}. All other datasets we use are licensed freely for academic use.

\subsection*{Annotation of \lmlml\ Dataset}
We conduct a user study to collect pairwise coherence judgments on our language model output dataset. As part of our crowd-sourced user study on Amazon Mechanical Turk to collect these coherence judgements, we do not collect any personal information from the participants. Based on the average time spent to perform the tasks, participants were paid the equivalent of 16 USD per hour for their work. The annotation instructions and interface provided to the participants are included in \Cref{subsec:appendix-user-study}. 

One potential issue is that the language model output that we generate from prompts may lead to malicious text generation by the models. We flagged the task to warn the workers that there may be potentially offensive content, and manually checked the final dataset post curation.

\subsection*{Applicability Across Languages}
All our experiments are conducted using data for the English language. However, as coherence and discourse relations in text are a universal concept, and our training data is automatically generated, we expect the permuted document task to be easily extensible to other languages.

%% file: files/appendix.tex
\subsection{\wsjl\ Permuted Document Task}
\label{subsec:appendix-wsj-example}

The examples for the permuted document task on the \wsjl\ data are shown in \Cref{tab:wsj_permute_eg}.

\subsection{Hard Negative Ranking Pseudocode}
\label{subsec:appendix-pseudocode}
The pseudocode for our hard negative mining through local sample ranking is given in \Cref{alg:hard-neg}.

\subsection{\lmlml\ User Study}
\label{subsec:appendix-user-study}

\begin{figure*}[h]
    \centering
    \includegraphics[scale=0.4]{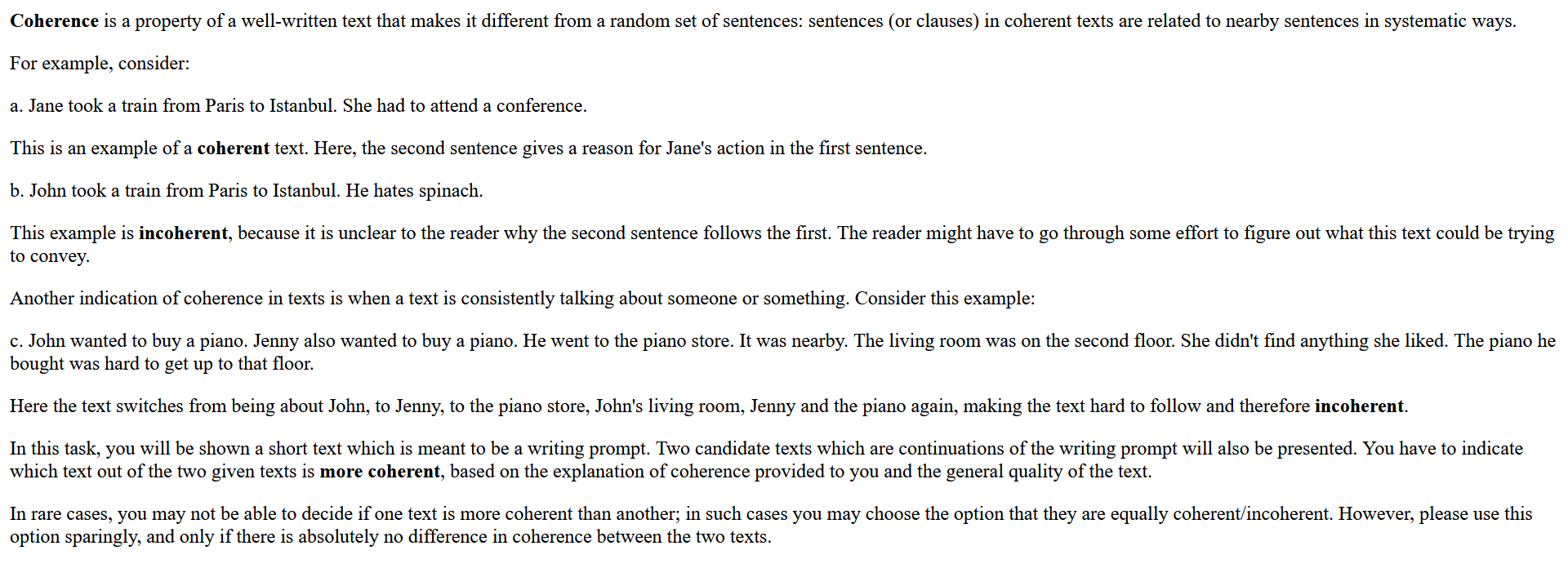}
    \includegraphics[scale=0.4]{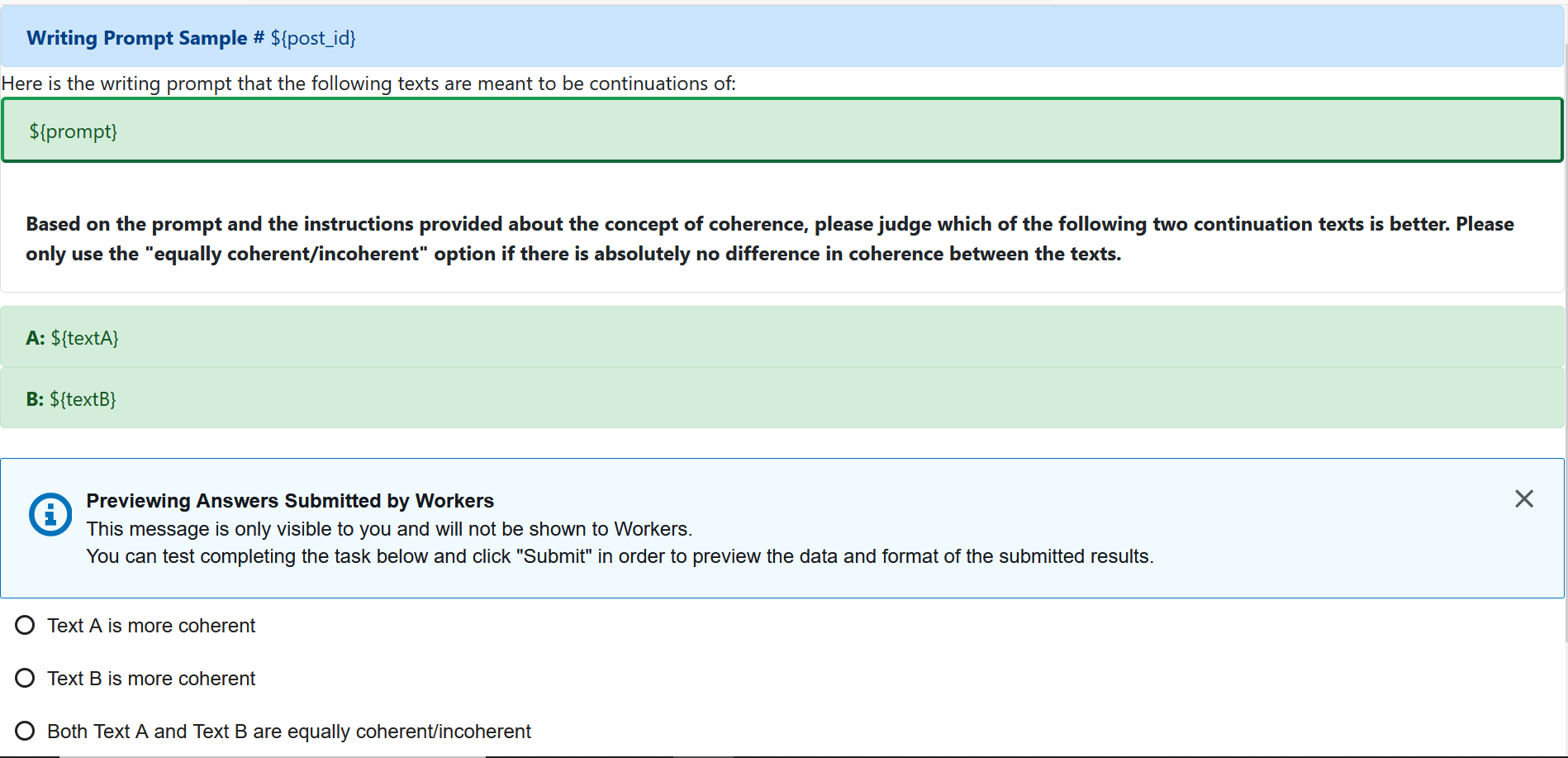}
    \caption{Instructions and study interface for the user study conducted on language model outputs.}
    \label{fig:lmlm_study}
\end{figure*}

The instructions and the interface provided to the workers in the user study comparing pairs of language model outputs is given in \Cref{fig:lmlm_study}. Workers were restricted to the native English speaking regions of Canada, United Kingdom and the United States and could only participate in our task if they had completed $>10,000$ HITs with a $>98\%$ acceptance rate. Each task was estimated to take 2 minutes, and workers were paid the equivalent of 16 USD per hour.

\subsection{Comparison of Existing State-of-The-Art Coherence Models}
\label{subsec:appendix-prev-evals}
We report the results obtained by  \citet{Pishdad2020HowCA} and \citet{rethinkingEACL} on their evaluation tasks for SOTA neural coherence models in \Cref{tab:prev_eval}. 

\subsection{Hyperparameters}
\label{subsec:appendix-hyperparams}

The hyperparameters used in our experiments are given in \Cref{tab:hyperparams}.

\begin{table}[t!]
\centering

\scalebox{0.80}{ \begin{tabular}{lc}  
\toprule
 \textbf{Parameters} & \textbf{Values} \\
\midrule
\multicolumn{2}{c}{\textbf{Margin-based Pairwise Ranking}} \\
\multicolumn{2}{c}{\textbf{(without XLnet fine-tuning)}} \\

\midrule

- margin & 0.1\\
- optimizer & AdamW \\
- scheduler & SWALR\\
- learning rate & 5e-6\\
- annealed to & 1e-6 \\
- anneal rate & 5000 steps \\
- batch-size & 1 \\
- XLNet model & base\\
- dimension size & 768\\

\midrule

\multicolumn{2}{c}{\textbf{Margin-based Pairwise Ranking}} \\
\midrule

- margin & 0.1\\
- optimizer & AdamW \\
- scheduler & SWALR\\
- learning rate & 5e-6\\
- annealed to & 1e-6 \\
- anneal rate & 5000 steps \\
- batch-size & 1 \\
- XLNet model & base\\
- dimension size & 768\\

\midrule

 \multicolumn{2}{c}{\textbf{Contrastive Learning}} \\
 \midrule
- margin & 0.1\\
- optimizer & AdamW \\
- scheduler & SWALR \\
- learning rate & 5e-6\\
- annealed to & 1e-6 \\
- anneal rate & 5000 steps \\
- batch-size & 1 \\
- XLNet model & base\\
- dimension size & 768\\

\midrule

\multicolumn{2}{c}{\textbf{Momentum Encoder with Hard Negative Mining}} \\
\midrule

- margin & 0.1\\
- optimizer & AdamW \\
- scheduler & SWALR \\
- learning rate & 5e-6\\
- annealed to & 1e-6 \\
- anneal rate & 1000 steps \\
- batch-size & 1 \\
- XLNet model & base\\
- dimension size & 768\\

\bottomrule
\end{tabular}}
\caption{Configuration parameters for training }
\label{tab:hyperparams}
\end{table}

\begin{table*}[t!]

    \footnotesize
    \centering
    \scalebox{0.88}{
    \begin{tabular}{p{0.02\textwidth} p{0.98\textwidth}}
    \toprule
         \multicolumn{2}{c}{\textbf{Original Document}} \\
         \midrule
          {\textbf{(S1)}}& Judy and I were in our back yard when the lawn started rolling like ocean waves. \\
         {\textbf{(S2)}} &We ran into the house to get Mame, but the next tremor threw me in the air and bounced me as I tried to get to my feet. \\
         {\textbf{(S3)}}& We are all fine here, although Mame was extremely freaked.\\
         {\textbf{(S4)}}& Books and tapes all over my room. \\
         {\textbf{(S5)}} &Not one thing in the house is where it is supposed to be, but the structure is fine.  \\
         \midrule
         \multicolumn{2}{c}{\textbf{Permuted Document}}  \\
         \midrule
           {\textbf{(S4)}}& Books and tapes all over my room. \\
           {\textbf{(S3)}}& We are all fine here, although Mame was extremely freaked. \\
           {\textbf{(S2)}}& We ran into the house to get Mame, but the next tremor threw me in the air and bounced me as I tried to get to my feet. \\
           {\textbf{(S5)}}& Not one thing in the house is where it is supposed to be, but the structure is fine.  \\
           {\textbf{(S1)}}& Judy and I were in our back yard when the lawn started rolling like ocean waves. \\

         \bottomrule
    \end{tabular}}
\caption{Examples showing the original coherent document and the incoherent document created by permuting the sentences of the original. Text taken from WSJ-1778.}
    
    \label{tab:wsj_permute_eg}

\end{table*}

\begin{algorithm*}[h]
\caption{Local Negative Sample Ranking}

\begin{algorithmic}[1]
\Require Training data $D$ in which each instance consists of a positive document and $h$ negative documents, model $\theta$
\State {\textsc Initialize} empty hard negative array $\hat D^-$ for each instance  $\in D$
\Procedure{HardNegativeRanking}{$\theta,D$}       
    \State Partition the dataset into sets of $x$ instances $D_{1} \dots D_{r}$
    
    \For{$i=1$ \ldots $r$}
        \If{i==0} 
        \Comment{\blue{No hard negatives for first iteration}}
            \For{$j=1 \ldots x$}
                \State Randomly sample $N$ negatives from $D^-_{(i,j)}$  and store in $\hat D^-_{(i,j)}$
            \EndFor
        \EndIf

    \State {\textsc Train} $\theta$ with ($D^+_i$, $\hat D^-_{i}$)

        \For{$j=1 \ldots x$}
        \State {\textsc Score} all the $h$ negative documents in $D^-_{(i+1,j)}$ 
        \State {\textsc Sort} $D^-_{(i+1,j)}$  in descending order of scores
        \State Get $N$ top scoring negative documents and store in $\hat D^-_{(i+1, j)}$
        \\
        \Comment{\blue{Store hard negatives for the next iteration}}
      
    \EndFor 
     \EndFor

\EndProcedure

\end{algorithmic}
\label{alg:hard-neg}
\end{algorithm*}

\begin{table*}[th]
    \centering
    \begin{tabular}{l|l|c|c}
    \toprule
    
        \multicolumn{4}{c}{As reported by \citet{Pishdad2020HowCA}} \\
        \midrule
        \textbf{Task} & \textbf{Dataset} & \unc\ & \citet{mesgar-strube-2018-neural} \\
        \midrule
        Permuted Document & Visual Storytelling & \textbf{88.42} & 82.25 \\
        Permuted Document & ROCStories & \textbf{94.80} & 89.55 \\
        Permuted Document & Dialogue  & \textbf{97.21} & 90.79 \\
        Permuted Document & HellaSwag & \textbf{83.92} & 69.38 \\
        Permuted Document & PDTB & \textbf{92.85} & 61.96 \\
        Connective Substitution & PDTB & \textbf{96.46} & 84.99 \\
        Topic Switching & Visual Storytelling & \textbf{92.10} & 64.81 \\
        Topic Switching & ROCStories & \textbf{94.62} & 67.85 \\
        Topic Switching & Dialogue  & \textbf{71.74} & 68.41 \\
        Topic Switching & PDTB & \textbf{70.89} & 52.33 \\
       \midrule 
        \multicolumn{4}{c}{As reported by \citet{rethinkingEACL}} \\
    \midrule 
       \textbf{Task} & \textbf{Dataset} & \unc\  & \lcd{}\\
       \midrule
        Permuted Document & \wsjl{}  & \textbf{93.19} & 91.77 \\
        Abstractive Summarization (Agr.)& CNN & \textbf{0.68} & 0.55 \\
        Extractive Summarization (Agr.)& DUC & 0.35 & \textbf{0.38}  \\

        Machine Translation (Agr.) & WMT & 0.77 & \textbf{0.78} \\
        (Trained) Machine Translation (Agr.)& WMT & \textbf{0.83} & 0.75  \\

      \bottomrule  
    \end{tabular}
    \caption{Results reported by \citet{rethinkingEACL}  and \citet{Pishdad2020HowCA} on various tasks and datasets that compare the \citet{unifiedcoherence} (\unc{}) model to two other SOTA neural coherence models proposed by \citet{mesgar-strube-2018-neural} and \citet{xu-etal-2019-cross} (\lcd{}). Except those marked by (Agr.) which report agreement with humans, all other tasks report accuracies. We only include tasks that directly test discourse coherence phenomena. }
    \label{tab:prev_eval}
\end{table*}

%% file: main.bbl
\begin{thebibliography}{51}
\expandafter\ifx\csname natexlab\endcsname\relax\def\natexlab#1{#1}\fi

\bibitem[{Arora et~al.(2019)Arora, Khandeparkar, Khodak, Plevrakis, and
  Saunshi}]{pmlr-v97-saunshi19a}
Sanjeev Arora, Hrishikesh Khandeparkar, Mikhail Khodak, Orestis Plevrakis, and
  Nikunj Saunshi. 2019.
\newblock \href {http://proceedings.mlr.press/v97/saunshi19a.html} {A
  theoretical analysis of contrastive unsupervised representation learning}.
\newblock In \emph{Proceedings of the 36th International Conference on Machine
  Learning}, volume~97 of \emph{Proceedings of Machine Learning Research},
  pages 5628--5637. PMLR.

\bibitem[{Barzilay and Lapata(2008)}]{Barzilay2008ModelingLC}
R.~Barzilay and Mirella Lapata. 2008.
\newblock Modeling local coherence: An entity-based approach.
\newblock \emph{Computational Linguistics}, 34:1--34.

\bibitem[{Barzilay and Lapata(2005)}]{Barzilay:2005}
Regina Barzilay and Mirella Lapata. 2005.
\newblock Modeling local coherence: An entity-based approach.
\newblock In \emph{Proceedings of the 43rd Annual Meeting on Association for
  Computational Linguistics}, ACL '05, pages 141--148, Ann Arbor, Michigan.
  Association for Computational Linguistics.

\bibitem[{Brown et~al.(2020)Brown, Mann, Ryder, Subbiah, Kaplan, Dhariwal,
  Neelakantan, Shyam, Sastry, Askell, Agarwal, Herbert-Voss, Kr{\"u}ger,
  Henighan, Child, Ramesh, Ziegler, Wu, Winter, Hesse, Chen, Sigler, Litwin,
  Gray, Chess, Clark, Berner, McCandlish, Radford, Sutskever, and
  Amodei}]{GPT-3}
T.~Brown, B.~Mann, Nick Ryder, Melanie Subbiah, J.~Kaplan, Prafulla Dhariwal,
  Arvind Neelakantan, Pranav Shyam, Girish Sastry, Amanda Askell, Sandhini
  Agarwal, Ariel Herbert-Voss, G.~Kr{\"u}ger, T.~Henighan, R.~Child, Aditya
  Ramesh, D.~Ziegler, Jeffrey Wu, Clemens Winter, Christopher Hesse, Mark Chen,
  E.~Sigler, Mateusz Litwin, Scott Gray, Benjamin Chess, J.~Clark, Christopher
  Berner, Sam McCandlish, A.~Radford, Ilya Sutskever, and Dario Amodei. 2020.
\newblock Language models are few-shot learners.
\newblock \emph{ArXiv}, abs/2005.14165.

\bibitem[{Celikyilmaz et~al.(2018)Celikyilmaz, Bosselut, He, and
  Choi}]{celikyilmaz-etal-2018-deep}
Asli Celikyilmaz, Antoine Bosselut, Xiaodong He, and Yejin Choi. 2018.
\newblock \href {https://doi.org/10.18653/v1/N18-1150} {Deep communicating
  agents for abstractive summarization}.
\newblock In \emph{Proceedings of the 2018 Conference of the North {A}merican
  Chapter of the Association for Computational Linguistics: Human Language
  Technologies, Volume 1 (Long Papers)}, pages 1662--1675, New Orleans,
  Louisiana. Association for Computational Linguistics.

\bibitem[{Chen et~al.(2020)Chen, Kornblith, Norouzi, and Hinton}]{Chen2020ASF}
Ting Chen, Simon Kornblith, Mohammad Norouzi, and Geoffrey~E. Hinton. 2020.
\newblock A simple framework for contrastive learning of visual
  representations.
\newblock \emph{ArXiv}, abs/2002.05709.

\bibitem[{Collobert et~al.(2011)Collobert, Weston, Bottou, Karlen, Kavukcuoglu,
  and Kuksa}]{collobert2011natural}
Ronan Collobert, Jason Weston, L{\'e}on Bottou, Michael Karlen, Koray
  Kavukcuoglu, and Pavel Kuksa. 2011.
\newblock Natural language processing (almost) from scratch.
\newblock \emph{The Journal of Machine Learning Research}, 12:2493--2537.

\bibitem[{Conneau et~al.(2017)Conneau, Kiela, Schwenk, Barrault, and
  Bordes}]{Conneau2017SupervisedLO}
Alexis Conneau, Douwe Kiela, Holger Schwenk, Lo{\"i}c Barrault, and Antoine
  Bordes. 2017.
\newblock Supervised learning of universal sentence representations from
  natural language inference data.
\newblock In \emph{EMNLP}.

\bibitem[{Dai et~al.(2019)Dai, Yang, Yang, Carbonell, Le, and
  Salakhutdinov}]{Dai2019TransformerXLAL}
Zihang Dai, Z.~Yang, Yiming Yang, J.~Carbonell, Quoc~V. Le, and
  R.~Salakhutdinov. 2019.
\newblock Transformer-xl: Attentive language models beyond a fixed-length
  context.
\newblock In \emph{ACL}.

\bibitem[{Elsner and Charniak(2011)}]{Elsner:2011}
Micha Elsner and Eugene Charniak. 2011.
\newblock Extending the entity grid with entity-specific features.
\newblock In \emph{Proceedings of the 49th Annual Meeting of the Association
  for Computational Linguistics: Human Language Technologies: Short Papers -
  Volume 2}, HLT '11, pages 125--129, Portland, Oregon. Association for
  Computational Linguistics.

\bibitem[{Fabbri et~al.(2020)Fabbri, Kry{\'s}ci{\'n}ski, McCann, Xiong, Socher,
  and Radev}]{summeval}
Alexander~R Fabbri, Wojciech Kry{\'s}ci{\'n}ski, Bryan McCann, Caiming Xiong,
  Richard Socher, and Dragomir Radev. 2020.
\newblock Summeval: Re-evaluating summarization evaluation.
\newblock \emph{arXiv preprint arXiv:2007.12626}.

\bibitem[{Fan et~al.(2018)Fan, Lewis, and Dauphin}]{WritingPrompts}
Angela Fan, Mike Lewis, and Yann Dauphin. 2018.
\newblock \href {https://doi.org/10.18653/v1/P18-1082} {Hierarchical neural
  story generation}.
\newblock In \emph{Proceedings of the 56th Annual Meeting of the Association
  for Computational Linguistics (Volume 1: Long Papers)}, pages 889--898,
  Melbourne, Australia. Association for Computational Linguistics.

\bibitem[{Feng and Hirst(2012)}]{Feng:2012}
Vanessa~Wei Feng and Graeme Hirst. 2012.
\newblock Extending the entity-based coherence model with multiple ranks.
\newblock In \emph{Proceedings of the 13th Conference of the European Chapter
  of the Association for Computational Linguistics}, EACL '12, pages 315--324,
  Avignon, France. Association for Computational Linguistics.

\bibitem[{Feng et~al.(2014)Feng, Lin, and Hirst}]{Feng:2014}
Vanessa~Wei Feng, Ziheng Lin, and Graeme Hirst. 2014.
\newblock The impact of deep hierarchical discourse structures in the
  evaluation of text coherence.
\newblock In \emph{COLING}.

\bibitem[{Grosz and Sidner(1986)}]{Grosz1986AttentionIA}
B.~Grosz and C.~Sidner. 1986.
\newblock Attention, intentions, and the structure of discourse.
\newblock \emph{Comput. Linguistics}, 12:175--204.

\bibitem[{Gutmann and Hyvärinen(2010)}]{pmlr-v9-gutmann10a}
Michael Gutmann and Aapo Hyvärinen. 2010.
\newblock \href {http://proceedings.mlr.press/v9/gutmann10a.html}
  {Noise-contrastive estimation: A new estimation principle for unnormalized
  statistical models}.
\newblock In \emph{Proceedings of the Thirteenth International Conference on
  Artificial Intelligence and Statistics}, volume~9 of \emph{Proceedings of
  Machine Learning Research}, pages 297--304, Chia Laguna Resort, Sardinia,
  Italy. PMLR.

\bibitem[{Halliday and Hasan(1976)}]{Halliday76}
Michael Halliday and Ruqaiya Hasan. 1976.
\newblock \emph{{Cohesion in English}}, chapter~xx. Longman, London.

\bibitem[{Hassan et~al.(2018)Hassan, Aue, Chen, Chowdhary, Clark, Federmann,
  Huang, Junczys-Dowmunt, Lewis, Li, Liu, Liu, Luo, Menezes, Qin, Seide, Tan,
  Tian, Wu, Wu, Xia, Zhang, Zhang, and Zhou}]{Hassan2018AchievingHP}
Hany Hassan, Anthony Aue, Chang Chen, Vishal Chowdhary, Jonathan~R. Clark,
  Christian Federmann, Xuedong Huang, Marcin Junczys-Dowmunt, William Lewis,
  Mu~Li, Shujie Liu, T.~M. Liu, Renqian Luo, Arul Menezes, Tao Qin, Frank
  Seide, Xu~Tan, Fei Tian, Lijun Wu, Shuangzhi Wu, Yingce Xia, Dongdong Zhang,
  Zhirui Zhang, and Ming Zhou. 2018.
\newblock Achieving human parity on automatic chinese to english news
  translation.
\newblock \emph{ArXiv}, abs/1803.05567.

\bibitem[{He et~al.(2019)He, Fan, Wu, Xie, and Girshick}]{he2019moco}
Kaiming He, Haoqi Fan, Yuxin Wu, Saining Xie, and Ross Girshick. 2019.
\newblock Momentum contrast for unsupervised visual representation learning.
\newblock \emph{arXiv preprint arXiv:1911.05722}.

\bibitem[{Hermann et~al.(2015)Hermann, Kocisk{\'y}, Grefenstette, Espeholt,
  Kay, Suleyman, and Blunsom}]{Hermann2015TeachingMT}
K.~Hermann, Tom{\'a}s Kocisk{\'y}, Edward Grefenstette, Lasse Espeholt, W.~Kay,
  Mustafa Suleyman, and P.~Blunsom. 2015.
\newblock Teaching machines to read and comprehend.
\newblock In \emph{NIPS}.

\bibitem[{Hosseini-Asl et~al.(2020)Hosseini-Asl, McCann, Wu, Yavuz, and
  Socher}]{hosseiniasl2020simple}
Ehsan Hosseini-Asl, Bryan McCann, Chien-Sheng Wu, Semih Yavuz, and Richard
  Socher. 2020.
\newblock \href {http://arxiv.org/abs/2005.00796} {A simple language model for
  task-oriented dialogue}.

\bibitem[{Huang et~al.(2020)Huang, Sharma, Sun, Xia, Zhang, Pronin,
  Padmanabhan, Ottaviano, and Yang}]{Huang20}
Jui-Ting Huang, Ashish Sharma, Shuying Sun, Li~Xia, David Zhang, Philip Pronin,
  Janani Padmanabhan, Giuseppe Ottaviano, and Linjun Yang. 2020.
\newblock \href {https://doi.org/10.1145/3394486.3403305} {Embedding-based
  retrieval in facebook search}.
\newblock In \emph{Proceedings of the 26th ACM SIGKDD International Conference
  on Knowledge Discovery \& Data Mining}, KDD '20, page 2553–2561, New York,
  NY, USA. Association for Computing Machinery.

\bibitem[{Karpukhin et~al.(2020)Karpukhin, Oğuz, Min, Lewis, Wu, Edunov, Chen,
  and tau Yih}]{Karpukhin2020DensePR}
Vladimir Karpukhin, Barlas Oğuz, Sewon Min, Patrick Lewis, Ledell~Yu Wu,
  Sergey Edunov, Danqi Chen, and Wen tau Yih. 2020.
\newblock Dense passage retrieval for open-domain question answering.
\newblock \emph{ArXiv}, abs/2004.04906.

\bibitem[{Krippendorff(2011)}]{Krippendorff2011ComputingKA}
K.~Krippendorff. 2011.
\newblock Computing krippendorff\'s alpha-reliability.

\bibitem[{L{\"a}ubli et~al.(2018)L{\"a}ubli, Sennrich, and
  Volk}]{Lubli2018HasMT}
Samuel L{\"a}ubli, Rico Sennrich, and Martin Volk. 2018.
\newblock Has machine translation achieved human parity? a case for
  document-level evaluation.
\newblock In \emph{EMNLP}.

\bibitem[{Li and Jurafsky(2017)}]{li-jurafsky:2017}
Jiwei Li and Dan Jurafsky. 2017.
\newblock Neural net models of open-domain discourse coherence.
\newblock In \emph{Proceedings of the 2017 Conference on Empirical Methods in
  Natural Language Processing}, pages 198--209, Copenhagen, Denmark.
  Association for Computational Linguistics.

\bibitem[{Lin et~al.(2011)Lin, Ng, and Kan}]{Lin:2011}
Ziheng Lin, Hwee~Tou Ng, and Min-Yen Kan. 2011.
\newblock Automatically evaluating text coherence using discourse relations.
\newblock In \emph{Proceedings of the 49th Annual Meeting of the Association
  for Computational Linguistics: Human Language Technologies - Volume 1}, HLT
  '11, pages 997--1006, Portland, Oregon. Association for Computational
  Linguistics.

\bibitem[{Liu et~al.(2017)Liu, Lu, Yang, Qu, Zhu, and Li}]{Liu2017GenerativeAN}
Linqing Liu, Yao Lu, Min Yang, Qiang Qu, Jia Zhu, and Hongyan Li. 2017.
\newblock Generative adversarial network for abstractive text summarization.
\newblock \emph{ArXiv}, abs/1711.09357.

\bibitem[{Mesgar and Strube(2018)}]{mesgar-strube-2018-neural}
Mohsen Mesgar and Michael Strube. 2018.
\newblock \href {https://www.aclweb.org/anthology/D18-1464} {A neural local
  coherence model for text quality assessment}.
\newblock In \emph{Proceedings of the 2018 Conference on Empirical Methods in
  Natural Language Processing}, pages 4328--4339, Brussels, Belgium.
  Association for Computational Linguistics.

\bibitem[{Mohiuddin et~al.(2018)Mohiuddin, Joty, and
  Tien~Nguyen}]{joty-etal-2018-coherence}
Muhammad~Tasnim Mohiuddin, Shafiq Joty, and Dat Tien~Nguyen. 2018.
\newblock \href {https://www.aclweb.org/anthology/P18-1052} {Coherence modeling
  of asynchronous conversations: A neural entity grid approach}.
\newblock In \emph{Proceedings of the 56th Annual Meeting of the Association
  for Computational Linguistics (Volume 1: Long Papers)}, pages 558--568,
  Melbourne, Australia. Association for Computational Linguistics.

\bibitem[{Mohiuddin et~al.(2021)Mohiuddin, Jwalapuram, Lin, and
  Joty}]{rethinkingEACL}
Tasnim Mohiuddin, Prathyusha Jwalapuram, Xiang Lin, and Shafiq Joty. 2021.
\newblock \href {https://www.aclweb.org/anthology/2021.eacl-main.308}
  {Rethinking coherence modeling: Synthetic vs. downstream tasks}.
\newblock In \emph{Proceedings of the 16th Conference of the European Chapter
  of the Association for Computational Linguistics: Main Volume}, pages
  3528--3539, Online. Association for Computational Linguistics.

\bibitem[{Moon et~al.(2019)Moon, Mohiuddin, Joty, and Chi}]{unifiedcoherence}
Han~Cheol Moon, Tasnim Mohiuddin, Shafiq~R. Joty, and Xiaofei Chi. 2019.
\newblock \href {https://www.aclweb.org/anthology/D19-1231.pdf} {A unified
  neural coherence model}.
\newblock \emph{Proceedings of the 2019 Conference on Empirical Methods in
  Natural Language Processing and the 9th International Joint Conference on
  Natural Language Processing}, pages 2262--2272.

\bibitem[{Nguyen and Joty(2017)}]{dat-joty:2017}
Dat Nguyen and Shafiq Joty. 2017.
\newblock \href {https://doi.org/10.18653/v1/P17-1121} {A neural local
  coherence model}.
\newblock In \emph{Proceedings of the 55th Annual Meeting of the Association
  for Computational Linguistics (Volume 1: Long Papers)}, pages 1320--1330.
  Association for Computational Linguistics.

\bibitem[{Paulus et~al.(2018)Paulus, Xiong, and Socher}]{Paulus2018ADR}
Romain Paulus, Caiming Xiong, and R.~Socher. 2018.
\newblock A deep reinforced model for abstractive summarization.
\newblock \emph{ArXiv}, abs/1705.04304.

\bibitem[{Pennington et~al.(2014)Pennington, Socher, and
  Manning}]{pennington2014glove}
Jeffrey Pennington, Richard Socher, and Christopher~D. Manning. 2014.
\newblock \href {http://www.aclweb.org/anthology/D14-1162} {Glove: Global
  vectors for word representation}.
\newblock In \emph{Empirical Methods in Natural Language Processing (EMNLP)},
  pages 1532--1543.

\bibitem[{Peters et~al.(2018)Peters, Neumann, Iyyer, Gardner, Clark, Lee, and
  Zettlemoyer}]{Peters:2018}
Matthew~E. Peters, Mark Neumann, Mohit Iyyer, Matt Gardner, Christopher Clark,
  Kenton Lee, and Luke Zettlemoyer. 2018.
\newblock Deep contextualized word representations.
\newblock In \emph{Proc. of NAACL}.

\bibitem[{Pishdad et~al.(2020)Pishdad, Fancellu, Zhang, and
  Fazly}]{Pishdad2020HowCA}
L.~Pishdad, Federico Fancellu, Ran Zhang, and A.~Fazly. 2020.
\newblock How coherent are neural models of coherence?
\newblock In \emph{COLING}.

\bibitem[{Popel et~al.(2020)Popel, Tomkov{\'a}, Tomek, Łukasz Kaiser,
  Uszkoreit, Bojar, and Žabokrtsk{\'y}}]{CUBBITT}
M.~Popel, M.~Tomkov{\'a}, J.~Tomek, Łukasz Kaiser, Jakob Uszkoreit, Ondrej
  Bojar, and Z.~Žabokrtsk{\'y}. 2020.
\newblock Transforming machine translation: a deep learning system reaches news
  translation quality comparable to human professionals.
\newblock \emph{Nature Communications}, 11.

\bibitem[{Radford et~al.(2019)Radford, Wu, Amodei, Amodei, Clark, Brundage, and
  Sutskever}]{GPT2-Blog}
Alec Radford, Jeffrey Wu, Dario Amodei, Daniela Amodei, Jack Clark, Miles
  Brundage, and Ilya Sutskever. 2019.
\newblock \href {https://openai.com/blog/better-language-models/} {Better
  language models and their implications}.
\newblock \emph{OpenAI Blog}.

\bibitem[{Reiter(2018)}]{reiter2018structured}
Ehud Reiter. 2018.
\newblock A structured review of the validity of {BLEU}.
\newblock \emph{Computational Linguistics}, 44(3):393--401.

\bibitem[{Sharma et~al.(2019)Sharma, Huang, Hu, and Wang}]{sharma2019entity}
Eva Sharma, Luyang Huang, Zhe Hu, and Lu~Wang. 2019.
\newblock An entity-driven framework for abstractive summarization.
\newblock In \emph{Proceedings of the 2019 Conference on Empirical Methods in
  Natural Language Processing and the 9th International Joint Conference on
  Natural Language Processing (EMNLP-IJCNLP)}, pages 3271--3282.

\bibitem[{Sharma et~al.(2018)Sharma, Allen, Bakhshandeh, and
  Mostafazadeh}]{StoryCloze}
Rishi Sharma, J.~Allen, Omid Bakhshandeh, and N.~Mostafazadeh. 2018.
\newblock Tackling the story ending biases in the story cloze test.
\newblock In \emph{ACL}.

\bibitem[{Shen et~al.(2021)Shen, Mistica, Salehi, Li, Baldwin, and
  Qi}]{ailishen2021}
Aili Shen, Meladel Mistica, Bahar Salehi, Hang Li, Timothy Baldwin, and
  Jianzhong Qi. 2021.
\newblock \href {https://doi.org/10.1162/tacl_a_00388} {{Evaluating Document
  Coherence Modeling}}.
\newblock \emph{Transactions of the Association for Computational Linguistics},
  9:621--640.

\bibitem[{van~den Oord et~al.(2018)van~den Oord, Li, and
  Vinyals}]{Oord2018RepresentationLW}
A{\"a}ron van~den Oord, Y.~Li, and Oriol Vinyals. 2018.
\newblock Representation learning with contrastive predictive coding.
\newblock \emph{ArXiv}, abs/1807.03748.

\bibitem[{Vaswani et~al.(2017)Vaswani, Shazeer, Parmar, Uszkoreit, Jones,
  Gomez, Kaiser, and Polosukhin}]{VaswaniNIPS2017}
Ashish Vaswani, Noam Shazeer, Niki Parmar, Jakob Uszkoreit, Llion Jones,
  Aidan~N Gomez, \L~ukasz Kaiser, and Illia Polosukhin. 2017.
\newblock \href
  {https://proceedings.neurips.cc/paper/2017/file/3f5ee243547dee91fbd053c1c4a845aa-Paper.pdf}
  {Attention is all you need}.
\newblock In \emph{Advances in Neural Information Processing Systems},
  volume~30. Curran Associates, Inc.

\bibitem[{Wu et~al.(2020)Wu, Zhuang, Mosse, Yamins, and Goodman}]{Wu2020OnMI}
M.~Wu, Chengxu Zhuang, M.~Mosse, D.~Yamins, and Noah~D. Goodman. 2020.
\newblock On mutual information in contrastive learning for visual
  representations.
\newblock \emph{ArXiv}, abs/2005.13149.

\bibitem[{Xiong et~al.(2020)Xiong, Xiong, Li, Tang, Liu, Bennett, Ahmed, and
  Overwijk}]{Xiong2021ApproximateNN}
Lee Xiong, Chenyan Xiong, Ye~Li, Kwok-Fung Tang, Jialin Liu, Paul~N. Bennett,
  Junaid Ahmed, and Arnold Overwijk. 2020.
\newblock Approximate nearest neighbor negative contrastive learning for dense
  text retrieval.
\newblock \emph{ICLR}, abs/2007.00808.

\bibitem[{Xu et~al.(2019)Xu, Saghir, Kang, Long, Bose, Cao, and
  Cheung}]{xu-etal-2019-cross}
Peng Xu, Hamidreza Saghir, Jin~Sung Kang, Teng Long, Avishek~Joey Bose,
  Yanshuai Cao, and Jackie Chi~Kit Cheung. 2019.
\newblock \href {https://doi.org/10.18653/v1/P19-1067} {A cross-domain
  transferable neural coherence model}.
\newblock In \emph{Proceedings of the 57th Annual Meeting of the Association
  for Computational Linguistics}, pages 678--687, Florence, Italy. Association
  for Computational Linguistics.

\bibitem[{Xuan et~al.(2020)Xuan, Stylianou, Liu, and Pless}]{Xuan2020HardNE}
Hong Xuan, Abby Stylianou, Xiaotong Liu, and Robert Pless. 2020.
\newblock Hard negative examples are hard, but useful.
\newblock In \emph{ECCV}.

\bibitem[{Yang et~al.(2019)Yang, Dai, Yang, Carbonell, Salakhutdinov, and
  Le}]{XLNet}
Z.~Yang, Zihang Dai, Yiming Yang, J.~Carbonell, R.~Salakhutdinov, and Quoc~V.
  Le. 2019.
\newblock Xlnet: Generalized autoregressive pretraining for language
  understanding.
\newblock In \emph{NeurIPS}.

\bibitem[{Zhang et~al.(2020)Zhang, Sun, Galley, Chen, Brockett, Gao, Gao, Liu,
  and Dolan}]{zhang2019dialogpt}
Yizhe Zhang, Siqi Sun, Michel Galley, Yen-Chun Chen, Chris Brockett, Xiang Gao,
  Jianfeng Gao, Jingjing Liu, and Bill Dolan. 2020.
\newblock Dialogpt: Large-scale generative pre-training for conversational
  response generation.
\newblock In \emph{ACL, system demonstration}.

\end{thebibliography}
